\documentclass[conference]{IEEEtran}
\usepackage{cite}
\usepackage{amsmath,amssymb,amsfonts}
\usepackage{algorithmic}
\usepackage{graphicx}
\usepackage{textcomp}
\usepackage{xcolor}
\usepackage{booktabs}  
\usepackage{boldline}
\usepackage{hyperref}
\usepackage{diagbox}
\usepackage{psfrag}

\usepackage{tikz}
\newcommand\copyrighttext{%
  \footnotesize $\copyright$ 2020 IEEE. Personal use of this material is permitted. Permission from IEEE must be obtained for all other uses, in any current or future media, including reprinting/republishing this material for advertising or promotional purposes, creating new collective works, for resale or redistribution to servers or lists, or reuse of any copyrighted component of this work in other works.}
\newcommand\copyrightnotice{%
\begin{tikzpicture}[remember picture,overlay]
\node[anchor=south,yshift=10pt] at (current page.south) {\fbox{\parbox{\dimexpr\textwidth-\fboxsep-\fboxrule\relax}{\copyrighttext}}};
\end{tikzpicture}%
}

\usepackage{amssymb}
\usepackage{listings}
\usepackage{multirow}
\usepackage{multicol}
\usepackage{listings}
\usepackage{color}
\usepackage[aboveskip=2pt]{subfig}
\usepackage{amsthm}
\usepackage{array}
\usepackage{threeparttable}
\usepackage{wrapfig}
\usepackage{caption}
\usepackage{subfig}
\usepackage{flushend}

\newcommand{\floor}[1]{\left\lfloor #1 \right\rfloor}
\newcommand{\ceil}[1]{\left\lceil #1 \right\rceil}

\newcolumntype{C}[1]{>{\raggedleft\arraybackslash}m{#1}}

\theoremstyle{definition}

\definecolor{mygreen}{rgb}{0,0.6,0}
\definecolor{mygray}{rgb}{0.5,0.5,0.5}
\definecolor{mymauve}{rgb}{0.58,0,0.82}

\usepackage[]{algorithm2e}
\def\BibTeX{{\rm B\kern-.05em{\sc i\kern-.025em b}\kern-.08em
    T\kern-.1667em\lower.7ex\hbox{E}\kern-.125emX}}
\begin{document}

\title{R-TOD: Real-Time Object Detector with Minimized End-to-End Delay for Autonomous Driving

}

\author{\IEEEauthorblockN{Wonseok Jang${}^{1}$, Hansaem Jeong${}^{1}$, Kyungtae Kang${}^{2}$, Nikil Dutt${}^{3}$, and Jong-Chan Kim${}^{1,4}$}
\IEEEauthorblockA{{${}^{1}$Graduate School of Automotive Engineering, Kookmin University, Korea}\\ 
{${}^{2}$Department of Computer Science and Engineering, Hanyang University, Korea}\\
{${}^{3}$Department of Computer Science, University of California, Irvine, USA}\\
{${}^{4}$Department of Automobile and IT Convergence, Kookmin University, Korea}\\
{\small xkqhdl@kookmin.ac.kr, saem28@kookmin.ac.kr, ktkang@hanyang.ac.kr, dutt@uci.edu, jongchank@kookmin.ac.kr}}
}

\maketitle

\begin{abstract}
For realizing safe autonomous driving, the end-to-end delays of real-time object detection systems should be thoroughly analyzed and minimized. However, despite recent development of neural networks with minimized inference delays, surprisingly little attention has been paid to their end-to-end delays from an object's appearance until its detection is reported. With this motivation, this paper aims to provide more comprehensive understanding of the end-to-end delay, through which precise best- and worst-case delay predictions are formulated, and three optimization methods are implemented: (i) on-demand capture, (ii) zero-slack pipeline, and (iii) contention-free pipeline. Our experimental results show a 76\% reduction in the end-to-end delay of Darknet YOLO (You Only Look Once) v3 (from 1070~ms to 261~ms), thereby demonstrating the great potential of exploiting the end-to-end delay analysis for autonomous driving. Furthermore, as we only modify the system architecture and do not change the neural network architecture itself, our approach incurs no penalty on the detection accuracy.
\end{abstract}


\copyrightnotice
\section{Introduction}
\label{sec:introduction}

For the realization of safe autonomous driving, a vehicle's camera-based object detection system should detect hazardous on-road obstacles as quickly as possible to reduce the risk of collision. Thus, delays between the appearance of an object and its detection should be thoroughly analyzed and optimized. Since most recent object detectors are based on {\em deep neural networks} (DNNs), lots of studies have attempted to reduce DNN {\em inference delays} by developing lightweight neural networks~\cite{han2015deep, howard2017mobilenets} or hardware acceleration techniques~\cite{jouppi2017datacenter, nvdla}. However, surprisingly little attention has been paid to the analysis and optimization of the {\em end-to-end delays} of real-time object detection systems, which include not only inference delays but also other pre- and post-processing delays such as resizing images and displaying results. In this paper, the end-to-end delay is defined by from the physical appearance of an object (e.g., change of a traffic signal or sudden pedestrian crossing) until its first detection is reported.

\begin{figure}
\centerline{\includegraphics[scale= 0.2]{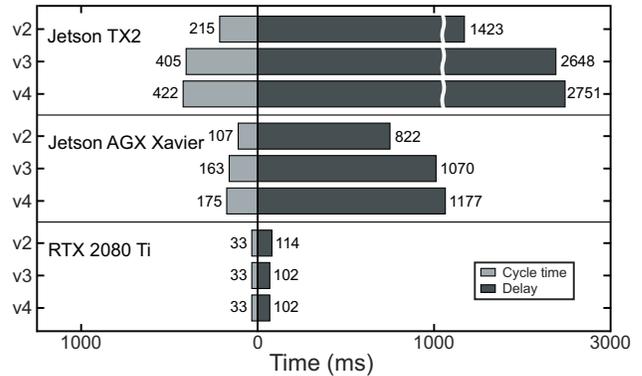}}
\caption{Average end-to-end delays and cycle times (i.e., inverse of frame rates) of Darknet YOLO (v2, v3, and v4) object detectors on three different hardware platforms.} 
\label{fig:pre_measure}
\vspace{-0.3cm}
\end{figure}

In autonomous driving systems, the frame rates, as well as the end-to-end delays, are commonly considered as the most important performance metrics~\cite{lin2018architectural}. In this regard, Fig.~\ref{fig:pre_measure} presents our preliminary measurement results of these metrics using three different versions of Darknet YOLO (You Only Look Once) object detectors~\cite{darknet, yolov2, redmon2018yolov3, bochkovskiy2020yolov4} on three Nvidia hardware platforms. For ease of comparison, cycle times (i.e., inverse of frame rates) are presented instead. The figure indicates that different combinations of DNNs and hardware platforms exhibit significantly varying performance. Another important observation highlighted here is the {\em imbalance between delays and cycle times}. For example, consider the case of Jetson AGX Xavier and YOLO v3: because its cycle time is 163~ms, we naively expect the delay to have a similar value; unexpectedly, the actual measured value is 1070~ms that is around six times the cycle time. It is apparent that the end-to-end delays are excessive in relation to the intrinsic cycle times, meaning that even if we are able to obtain object detection results with sufficient frequency, they still may suffer from significant time lags to the real world. Moreover, we found similar trends in other state-of-the-art object detection systems including Detectron2~\cite{wu2019detectron2} and Darkflow~\cite{trieu2018darkflow}. 
With this observation, we pose the following research questions:
\begin{itemize}
    \item What is the reason for the serious time lags compared with the cycle times of object detection systems?
    \item How can we minimize the end-to-end delays of object detection systems, and by how much?
\end{itemize}

To answer the first question, we thoroughly analyzed the internal architecture of the Darknet YOLO object detection system and measured every component that contributed to the end-to-end delay on our experimental platforms. From this investigation, we found that (i) Darknet YOLO relies on the OpenCV library for capturing real-time images from cameras, which manages a queue for buffering image frames; and (ii) it employs a three-stage multithreaded pipeline architecture for exploiting its {\em pipeline parallelism}~\cite{navarro2009analytical,gordon2006exploiting}. After carefully analyzing the internal architecture of Darknet YOLO, we have concluded that the reason behind the serious time lags is threefold: (i) there is an unnecessary queuing delay caused by the difference between the frame rate of the camera and that of the object detector; (ii) the lengths of the three pipeline stages are much too different causing significant idle times for shorter ones; and (iii) concurrently running GPU (Graphics Processing Unit) kernels and CPU (Central Processing Unit) threads incurs significant contention for shared memory bandwidth, leading to increased execution times, especially when an integrated GPU is used.

To answer the second question, while minimizing the end-to-end delays of the Darknet YOLO object detection system, we developed the following three techniques and evaluated them on an Nvidia Jetson AGX Xavier. Note that the numbers below are for YOLO v3 in particular:\\
{\bf (i) On-demand capture.} We removed unnecessary queuing delays by applying an {\em on-demand capture} method that does not use a queue between cameras and object detectors. Instead, an image frame is captured only when it is needed, thereby reducing the average delay from 1070~ms to 383~ms (i.e., a 64\% reduction).\\
{\bf (ii) Zero-slack pipeline.} We minimized pipeline idle times by postponing the beginning of the first pipeline stage as much as possible within its pipeline cycle such that the next cycle begins right after the end of the first stage. On average, this {\em zero-slack pipeline} saved an extra 73~ms, and together with on-demand capture, it reduced the overall delay to 310~ms. \\
{\bf (iii) Contention-free pipeline.} To minimize the memory bandwidth contention between concurrently running GPU kernels and CPU threads, we applied synchronized executions between them, thereby achieving temporal isolation of GPU kernel executions from CPU workloads. By applying this {\em contention-free pipeline}, the delay was even further reduced to 261~ms, which is about a 76\% reduction compared with the original one. Furthermore, to evaluate the tail behavior of the optimized system, the 99th percentile delays were compared, which showed an end-to-end delay reduction of about 67\%. 

The contributions of this study can be summarized as:
\begin{itemize}
    \item We present an in-depth analysis of the Darknet YOLO object detection system and our end-to-end delay analysis framework, and identify three problems with significant adverse impacts on its end-to-end delays.
    \item We propose a modified OpenCV library that supports {\em on-demand capture} that significantly reduces the end-to-end delays of various state-of-the-art object detectors without any executable file modification.
    \item We present {\em zero-slack} and {\em contention-free} pipeline architectures that alleviate the delays caused by imbalanced pipeline stages and memory bandwidth contention, respectively.
    \item The above techniques are implemented and evaluated on an Nvidia Jetson AGX Xavier CPU-GPU heterogeneous computing platform that shows 76\% average and 67\% 99th percentile delay reductions for YOLO v3, with no loss in detection quality.
\end{itemize}


\section{Background and Problem Description}
\label{sec:background}

\begin{figure}
\centerline{\includegraphics[scale= 0.6]{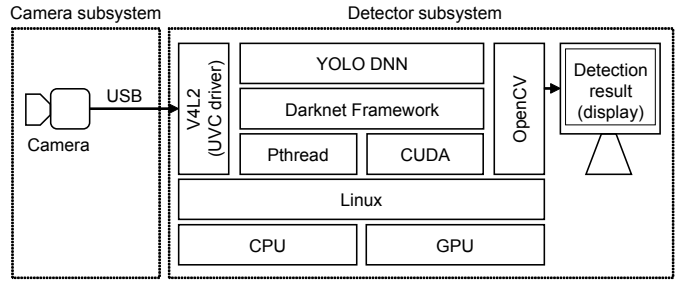}}
\caption{Overall hardware-software architecture.}
\vspace{-0.4cm}
\label{fig:overall_architecture}
\end{figure}

\subsection{Overall Hardware-Software Architecture}
\label{sec:overall}
Fig.~\ref{fig:overall_architecture} depicts our reference hardware-software architecture considered for object detection systems, which consists of (i) a camera subsystem and (ii) a detector subsystem. In the camera subsystem, we consider a single camera that is directly attached to the detector subsystem using a USB (Universal Serial Bus) connector. For the detector subsystem, a CPU-GPU heterogeneous computing platform is used. The Linux operating system is used to run the Darknet neural network framework based on the pthread library and the CUDA framework for managing CPU threads and GPU kernels, respectively. In the Darknet framework, we use pre-trained YOLO v2, v3, and v4 DNNs. As an interface between the two subsystems, the Video For Linux Two (V4L2)-based USB camera driver is used. To visualize object detection results, the OpenCV library is used to show bounding boxes around the detected objects. The remainder of this section briefly summarizes details regarding Darknet, YOLO, V4L2, and OpenCV.

\begin{figure*}
\psfrag{AX}{$\mathcal{A}$}
\psfrag{BB}{$\mathcal{S}$}
\centerline{\includegraphics[scale=0.64]{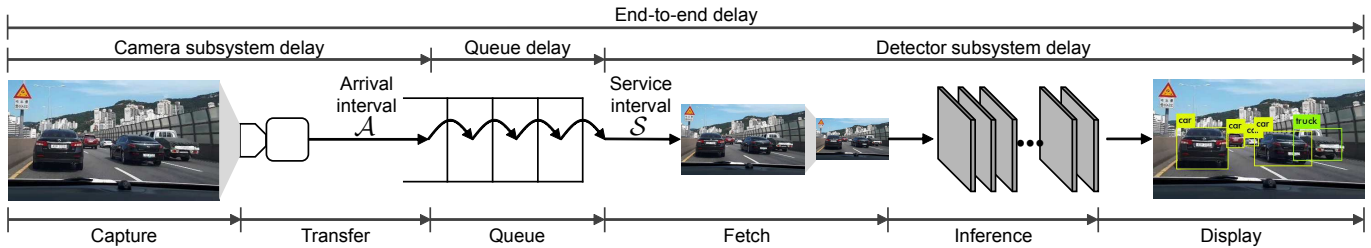}}
\caption{Overall flow and associated delays in an object detection system.}
\label{fig:five_stages}
\vspace{-0.5cm}
\end{figure*}

{\bf Darknet}~\cite{darknet} is an open source neural network framework developed as the baseline software architecture for the YOLO object detector. Darknet can be used for training a custom neural network as well as for an inference engine with pre-trained neural networks. Darknet can utilize GPUs as well as CPUs when running neural networks. Because it is written purely in C and CUDA with minimal dependencies, it is easy to port from a high-end workstation to a lightweight embedded system. Darknet supports the most commonly used neural network operators, such as convolution, pooling, and various activation functions. Thus, a neural network configuration file with layer-by-layer descriptions and a weights file are sufficient when implementing a neural network inference system.

{\bf YOLO} is one of the most famous single-stage object detectors based on the {\em convolutional neural network} (CNN) architecture. Because of its high speed and accuracy, it is widely used for implementing real-time object detection systems for autonomous driving~\cite{lin2018architectural, kato2018autoware, alcon2020timing}. YOLO has evolved from its original version~\cite{Redmon_2016_CVPR} to more recent versions (v2, v3, and v4)~\cite{yolov2, redmon2018yolov3, bochkovskiy2020yolov4} in which its object detection accuracy has been continually increased. Pre-trained neural network configuration files and weights files are available for each version; these are based on open datasets such as PASCAL VOC~\cite{Everingham15} and MS COCO~\cite{10.1007/978-3-319-10602-1_48}.

{\bf V4L2}~\cite{v4l2} is a device driver framework used to process real-time video images in Linux. For example, most USB cameras are supported by the UVC (USB Video Class) driver that relies on the V4L2 framework, especially when using its userspace application programming interface (API). The V4L2 API represents a camera as a character device file such that userspace applications can open it to fetch a sequence of images through the ioctl system call. Furthermore, users using the V4L2 API can also control camera parameters including the frame rate, pixel format, and exposure.

{\bf OpenCV}~\cite{opencv_library} is an open source computer vision library that provides various off-the-shelf computer vision algorithms. Recently, OpenCV has been used as the common infrastructure for various computer vision applications in the industry. In object detection systems, it is typically used for visualizing object detection results by drawing bounding boxes around each detected object and displaying the resulting images. Further, most object detectors, including Darknet, use the OpenCV C++ wrapper library, known as {\em VideoCapture}, instead of directly using the V4L2 API for capturing camera images.

\subsection{Problem Description}
\label{sec:problem}
Based on the hardware-software architecture described above, this study has two objectives: (i) to derive a timing model of object detection systems by analyzing the internal architecture of Darknet YOLO (Section~\ref{sec:delay_analysis}) and (ii) to present an optimal application architecture for object detection systems in terms of minimizing end-to-end delays (Section~\ref{sec:optimization}) and evaluate it with the Darknet YOLO object detection system (Section~\ref{sec:experiments}).

\section{End-to-End Delay Analysis}
\label{sec:delay_analysis}
{\bf Note:} Our notational conventions are as follows: (i) Roman capitals for constants (e.g., $F$), (ii) calligraphic capitals for probability distributions (e.g., $\mathcal{A}$), (iii) Roman lower-case letters for random variables (e.g., $d_{capt}$), and (iv) Roman lower-case letters with the superscripts min or max for the minimums or maximums of random variables (e.g., $d_{capt}^{max})$.

\subsection{Holistic Approach for the End-to-End Delay Analysis}
\label{sec:holistic}
Fig.~\ref{fig:five_stages} shows the end-to-end flow of data in an object detection system, along with the associated delays, that comprises a camera subsystem, a detector subsystem, and a V4L2 camera driver queue (henceforth simply referred to as ``queue'') for buffering images. The queue is located inside the detector subsystem; however, for ease of explanation, we will regard it as being between the two subsystems. This entire system can be modeled as a single-server queue of size $Q$. Throughout the analysis, we use the distribution $\mathcal{A}$ of image arrival intervals at the queue and the distribution $\mathcal{S}$ of object detection service intervals, where the service interval is defined as the time between two consecutive requests for retrieving images from the queue (i.e., beginning of object detection cycles) by the detector subsystem. These distributions are discussed later in this section.

 Our analysis of the end-to-end delay can now be summarized as follows:
\begin{itemize}
    \item {\bf Camera subsystem.} It incurs a delay defined by the time from the first appearance of an object to the queuing of its first image. We examine the components of this delay to formulate $\mathcal{A}$.
    \item {\bf Detector subsystem.} It incurs a delay defined by the time from the retrieval of an image from the queue to the appearance of the detection result on the display. We examine the components of this delay to formulate $\mathcal{S}$.
    \item {\bf Queue.} It incurs a delay that can be determined from $\mathcal{A}$ and $\mathcal{S}$ for a given queue size $Q$.
\end{itemize}
In Fig.~\ref{fig:five_stages}, six different delay components are present: (i) capture, (ii) transfer, (iii) queue, (iv) fetch, (v) inference, and (vi) display; these account for the end-to-end delay. Note that the inference delay is only one among many delays, thus necessitating a holistic approach for analyzing and minimizing end-to-end delays. Moreover, although these delay components appear to be isolated from each other, some are interrelated, further indicating the need for a holistic approach.

\subsection{Camera Subsystem}
\label{sec:camera_delay}
Images are captured by the camera (i.e., capture instant) at a frame rate of $F$, so the camera's {\em cycle time} $C$ between captures is $1/F$. At first glance, a newly appeared object seems to have to wait at most $C$ for the next capture instant. However, the object may experience a much longer delay before it is captured, that is, the {\em capture delay} denoted by $d_{capt}$. Because it is intertwined with the detector subsystem's service intervals, an exact formulation will be presented later in this section.

The amount of data in an image frame is determined by its resolution ($X \times Y$) and pixel format. Non-compressing pixel formats such as YUYV and RGBA have a fixed bits-per-pixel (bpp) value $P$. In contrast, compressing pixel formats such as H264 and MJPEG may vary the frame size depending on the image content. We assume non-compressing pixel formats because compressing pixel formats are non-deterministic and lossy, and they require an additional decryption process. Thus, assuming a fixed bpp, an image frame's size, denoted by $I$, can be calculated as $I$ = $(X \cdot Y \cdot P) / 8$ bytes.

\begin{table}
  \caption{USB bandwidth reservations.}
  \label{tab:USBband}
\centering
\begin{tabular}{ccccc}
\toprule
 Resolution&  $F$ & $B$ & $M$& Reserved (Required)\\
 (X $\times$ Y)&  (fps)& (bytes)& ($\mu$frames)& bandwidth (MB/s)\\
\toprule
  \multirow{2}{*}{320 $\times$ 240}&  20& 512& 32& 4 (3)\\
 &  30& 800& 32& 6 (4)\\
  \toprule
  \multirow{2}{*}{640 $\times$ 480}&  20& 1984& 32& 15 (12) \\
&  30& 2688& 32& 21 (18)\\

\toprule
\end{tabular}
\label{tab:bandwidth}
\vspace{-0.5cm}
\end{table}

Each captured image is transferred to the detector subsystem through a USB connection in the isochronous transfer mode, which provides a known latency by bandwidth reservation~\cite{lee2007handbook}. The camera driver interacts with the camera to determine the available bandwidth, which is expressed as the number of bytes, denoted by $B$, available during each USB microframe\footnote{USB manages time in units called ``frames" of 1 ms. At a high speed, each 1 ms frame is divided into eight ``microframes'' of 125~$\mu$s.} of length $U$=125~$\mu$s. A USB driver uses a buffer to group a {\em USB request block} (URB) of $M$ microframes (i.e., handling granularity). Table~\ref{tab:bandwidth} lists the example bandwidth reservations measured in our experimental platform using the YUYV pixel format with $P$ = 16.

Next, consider the {\em transfer delay}, denoted by $d_{tran}$, which is the time between the capture of an image and its addition to the queue. This can be estimated by counting the number of required microframes from given $I$, $B$, and $U$, as follows:
\begin{equation}
    d_{tran} = \left(\ceil{\frac{I}{B}} + 2\right) U + j,
    \label{eq:d_tran}
\end{equation}
where $j$ is the jitter caused by (i) waiting for the start of the next microframe after a capture ($U$ in the worst case), and (ii) waiting for the current URB to be grouped ($(M-1) \cdot U$ in the worst case). We denote the former as {\em microframe jitter} and the latter as {\em URB buffering jitter}, respectively. By summing the two, $j$ can have values in the range $0 \le j < M \cdot U$. Note that the constant 2 represents extra microframes containing protocol overheads.

\begin{figure}
\centerline{\includegraphics[scale=0.65]{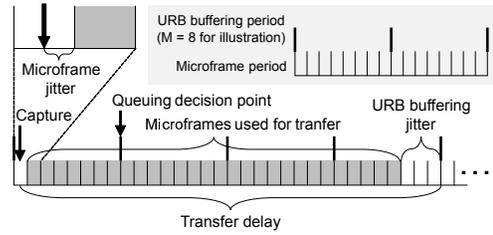}}
\caption{Transfer delay.}
\label{fig:tran}
\vspace{-0.7cm}
\end{figure}

Fig.~\ref{fig:tran} illustrates the transfer delay while taking this jitter into account. It indicates that the transfer delay is determined as the multiples of a URB buffering period of $M$ microframes, because the camera driver only checks image arrivals at that interval. Note that $M \cdot U$ = $32 \cdot 125~\textrm{$\mu$s}$ = 4~ms is true for every row of Table~\ref{tab:bandwidth}, although it can vary for different cameras.

\begin{figure}
\begin{center}
\subfloat[20 fps]{\label{fig:arrival_20fps}\includegraphics[width=.25\textwidth]{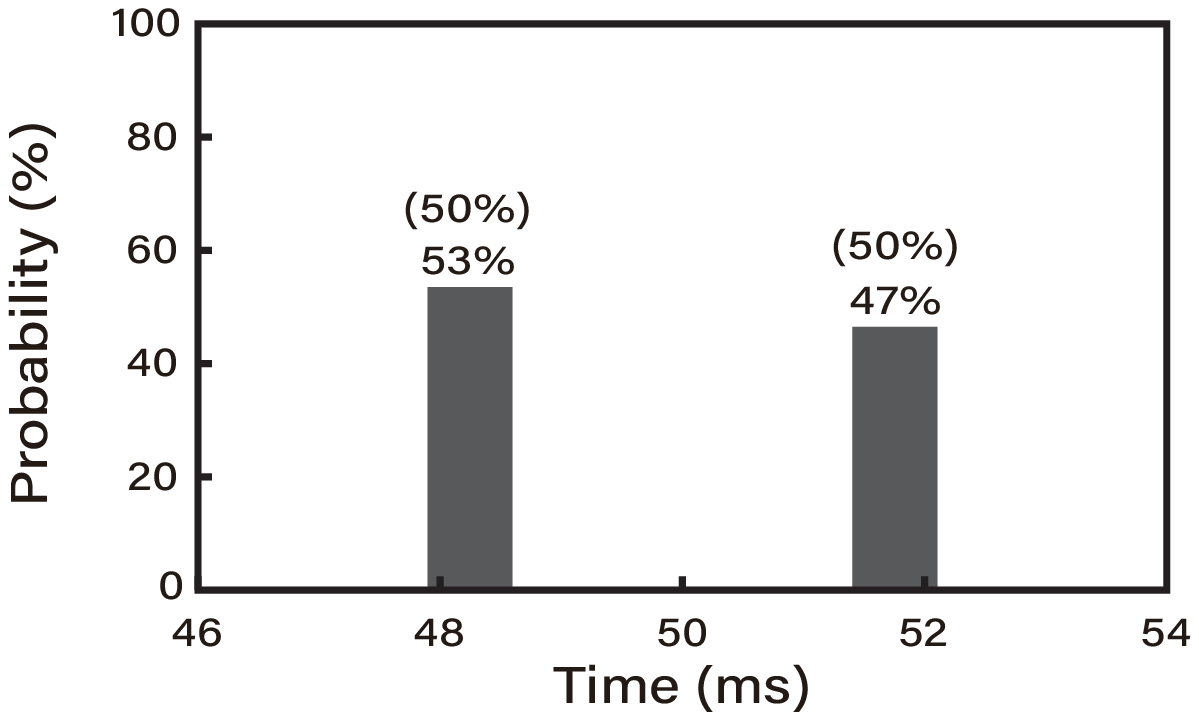}}
\subfloat[30 fps]{\label{fig:arrival_30fps}\includegraphics[width=.25\textwidth]{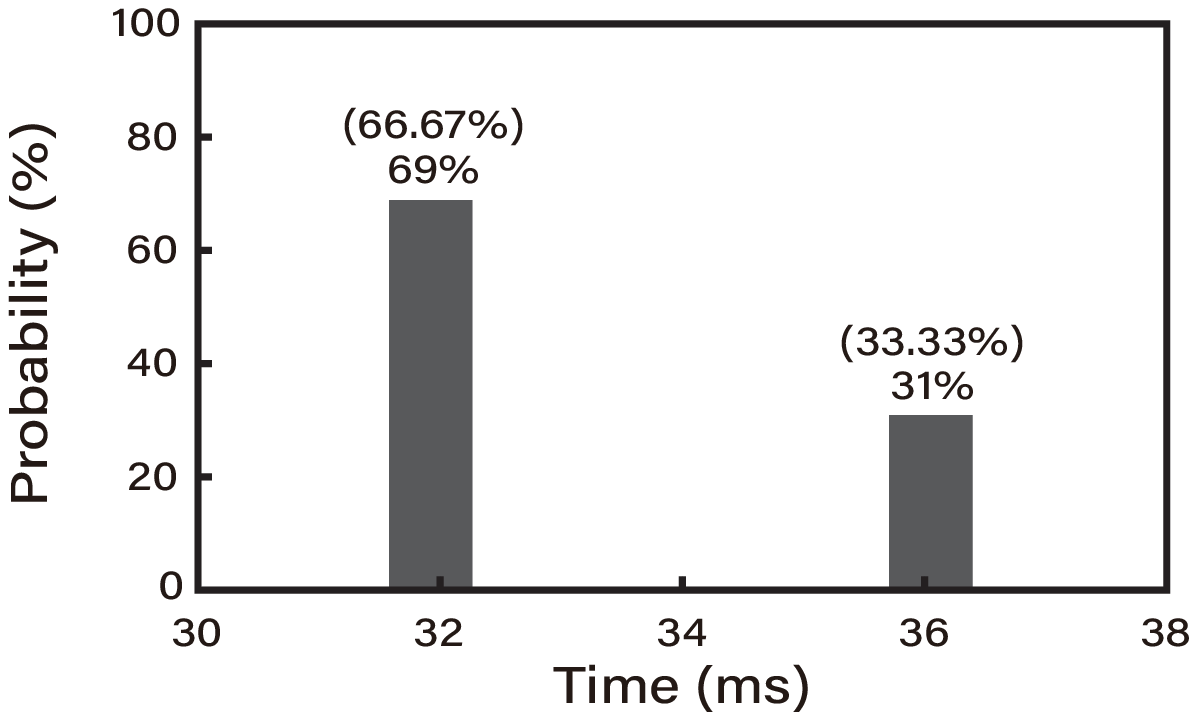}}
\caption{Image arrival interval distribution ($\mathcal{A})$ with varying camera frame rates. Probabilities in parentheses are estimates; measured probabilities are written below the estimates.}
\label{fig:arrivals}
\vspace{-0.6cm}
\end{center}
\end{figure}

An important consequence of this approach based on URB buffering is that the image arrival intervals are adjusted to the multiples of the {\em URB buffering period} ($M \cdot U$) (note that their expectation approximates to the constant camera cycle time $C$). Thus, $\mathcal{A}$ becomes a discrete probability distribution with (usually) two values $a^{min}$ and $a^{max}$, and the probability that the image arrival interval $a$ equals one of these values is, respectively, defined as follows:
\begin{equation}
\begin{centering}
\begin{aligned}
    p_1 = \textrm{Pr}\left\{a = a^{min}\right\}\textrm{ and } p_2 = \textrm{Pr}\left\{a = a^{max}\right\},
\end{aligned}
\end{centering}
\end{equation}
where
\begin{equation*}
\begin{centering}
\begin{aligned}
    a^{min} = \floor{\frac{C}{M \cdot U}} (M \cdot U) \textrm{ and }
    a^{max} = \ceil{\frac{C}{M \cdot U}} (M \cdot U),
\end{aligned}
\end{centering}
\end{equation*}
which represent the closest multiple of $M \cdot U$ to the floor and ceiling of $C$, respectively. The expected value of $a$ $(E(a))$ can then be approximated to $C$, i.e., $a^{min}p_1 + a^{max}p_2 \approx C$. Since $p_1 + p_2 = 1$, the probability distribution can be derived from given $C$, $M$, and $U$. Fig.~\ref{fig:arrivals} shows the measured arrival interval distributions at two different camera frame rates: 20 fps and 30 fps. Note that the results closely match our estimates, with minor discrepancies caused by limitations in the camera hardware with respect to the exact frame rate.

When an image frame arrives at the queue, it is either accepted by the queue or dropped, where some droppings are inevitable in case the image arrivals are faster than the retrievals. In reality, the queuing decision is made even before it fully arrives. From our protocol analysis, we found that the decision is made when the first URB buffering period containing the image frame ends, and we call the instant when the decision is made a {\em queuing decision point}, shown in Fig.~\ref{fig:tran}. It is apparent that we have a random time interval between the image capture and the queuing decision, which is at most a URB buffering period $(M \cdot U)$. We call this {\em queuing decision jitter}.

We revisit an object's capture delay $d_{capt}$, defined as the time lapse between the physical appearance of an object and its successful capture. A successful capture means that the object is queued after its transmission. A capture can be unsuccessful when it is later dropped by the queue, because the queue does not have sufficient room to accommodate the frame associate with the object. It is apparent that $d_{capt}$ is intertwined with the detector subsystem's service interval distribution $\mathcal{S}$. At the beginning of each object detection cycle, an image frame is retrieved from the queue, making room for a new image frame. The queuing decision point now plays a key role in enqueuing the frame; the decision point must occur after this moment for the frame associated with it to be accepted by the queue.

In the best case, $d_{capt}$ = 0 when an object appears right before a capture instant\footnote{In this paper, we do not consider the minimum exposure requirement of the camera. However, it can simply be assigned a value greater than zero as the minimum capture delay.} and the queue has sufficient room to accommodate the image frame associated with the object. In contrast, the worst case occurs when an object appears right after the capture instant that makes the queue just become full so that all subsequently captured images should be dropped by the queue until the queue becomes available. The worst-case $d_{capt}$ can now be drawn using the maximum number of image frames subsequently dropped within the longest object detection service interval\footnote{During each object detection cycle, three sequential modules (fetch, inference, and display, which will be explained in the following section) are executed in a pipeline manner to detect the object. Hence, this interval is determined by the maximum among the delays incurred by these modules.}, denoted by $s^{max}$. We can estimate this number by counting the number of queuing decision points within $s^{max}$, by taking the effect of the maximum queuing decision jitter ($M\cdot U$) into account. The time interval determined by the number of queuing decision points can be shorter than the interval obtained by the same number of capture instants by at most $M \cdot U$. From this observation, $s^{max}$ should be added by $M\cdot U$ to account for the maximum queuing decision jitter when it is divided by $C$, in order to obtain $d_{capt}$, which yields the range of $d_{capt}$, as follows:
\begin{equation}
    0 \le d_{capt} < \ceil{\frac{s^{max} + M \cdot U}{C}}C.
    \label{eq:d_capt}
\end{equation}

By summing Eq.~\eqref{eq:d_tran} and Eq.~\eqref{eq:d_capt}, we have the minimum and maximum camera subsystem delays, denoted by $d_{camera}^{min}$ and $d_{camera}^{max}$, respectively, as follows:
\begin{equation}
\begin{centering}
\begin{aligned}
    &d_{camera}^{min} = \left(\ceil{\frac{I}{B}} + 2\right) U\textrm{ and}\\
    &d_{camera}^{max} = \ceil{\frac{s^{max} + M \cdot U}{C}}C +\left(\ceil{\frac{I}{B}}
    + 2 + M\right) U.
\end{aligned}
\end{centering}
\label{eq:d_camera_final}
\end{equation}

\subsection{Detector Subsystem}
\label{sec:detector}
The detector subsystem comprises three modules: (i) fetch, (ii) inference, and (iii) display. We now introduce these modules along with the delay components associated with them. The detector subsystem's service interval distribution $\mathcal{S}$ will be identified after analyzing and profiling the three modules on an Nvidia Jetson AGX Xavier. For now, no other background workload is assumed.

\begin{figure}
\centering
\subfloat[$\mathcal{D}_{fetch}(r_{in}, r_{out})$]{\label{fig:prof_f}\includegraphics[width=.23\textwidth]{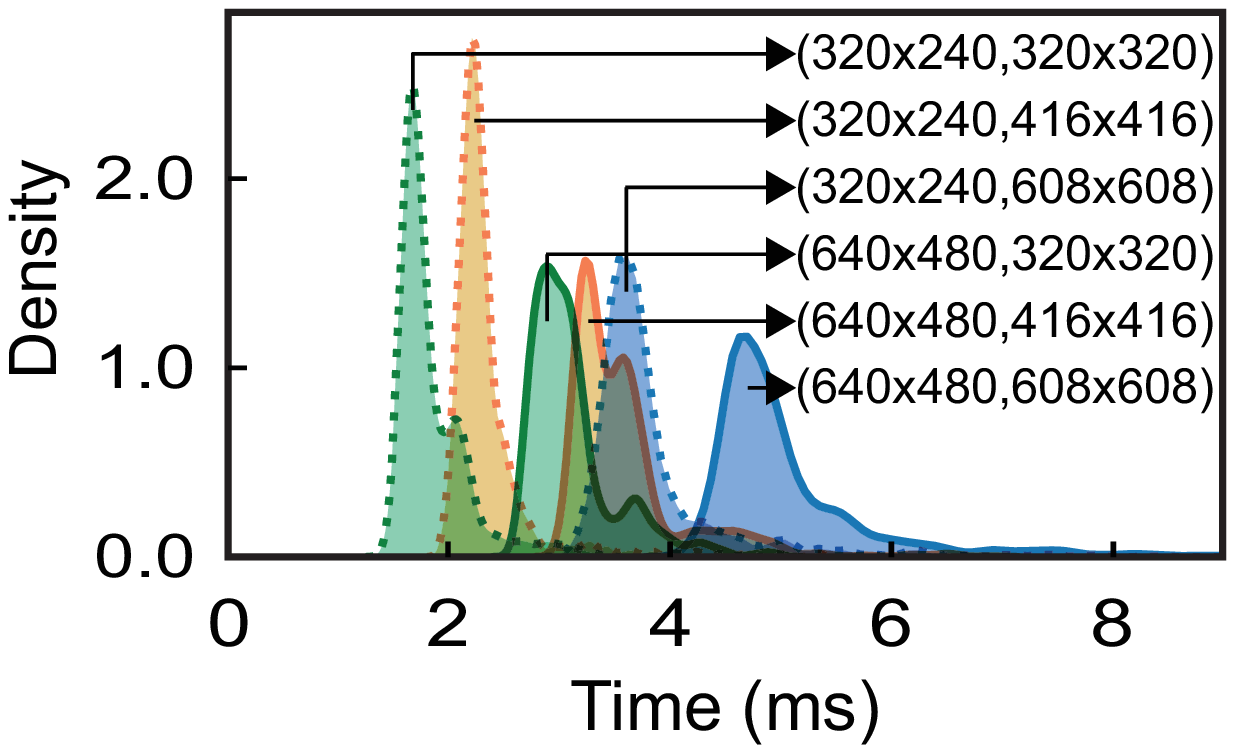}}
\quad
\subfloat[$e_{fetch}$, $b_{fetch}$, and $d_{fetch}$]{\label{fig:prof_f2}\includegraphics[width=.23\textwidth]{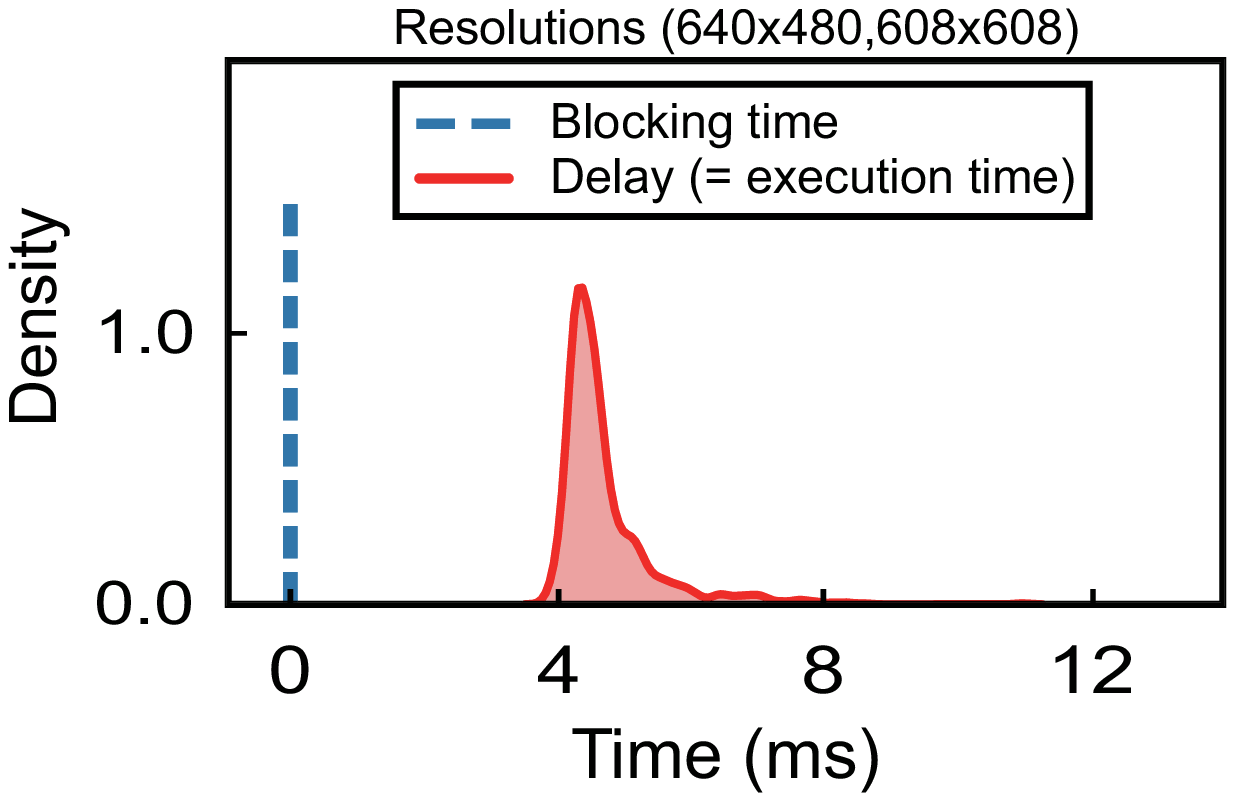}}
\\
\subfloat[$\mathcal{D}_{infer}(r_{nn}$)]{\label{fig:prof_i}\includegraphics[width=.23\textwidth]{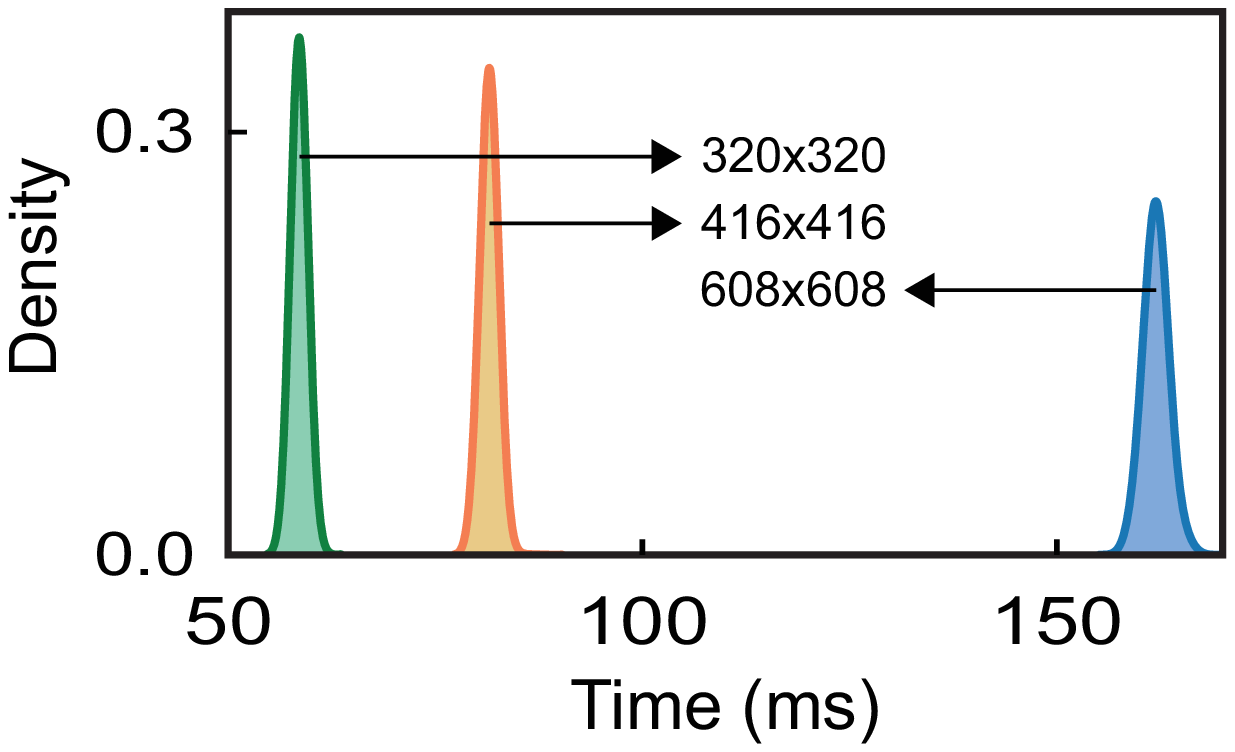}}
\quad
\subfloat[$e_{infer}^{CPU}$, $e_{infer}^{GPU}$, and $d_{infer}$]{\label{fig:prof_f3}\includegraphics[width=.23\textwidth]{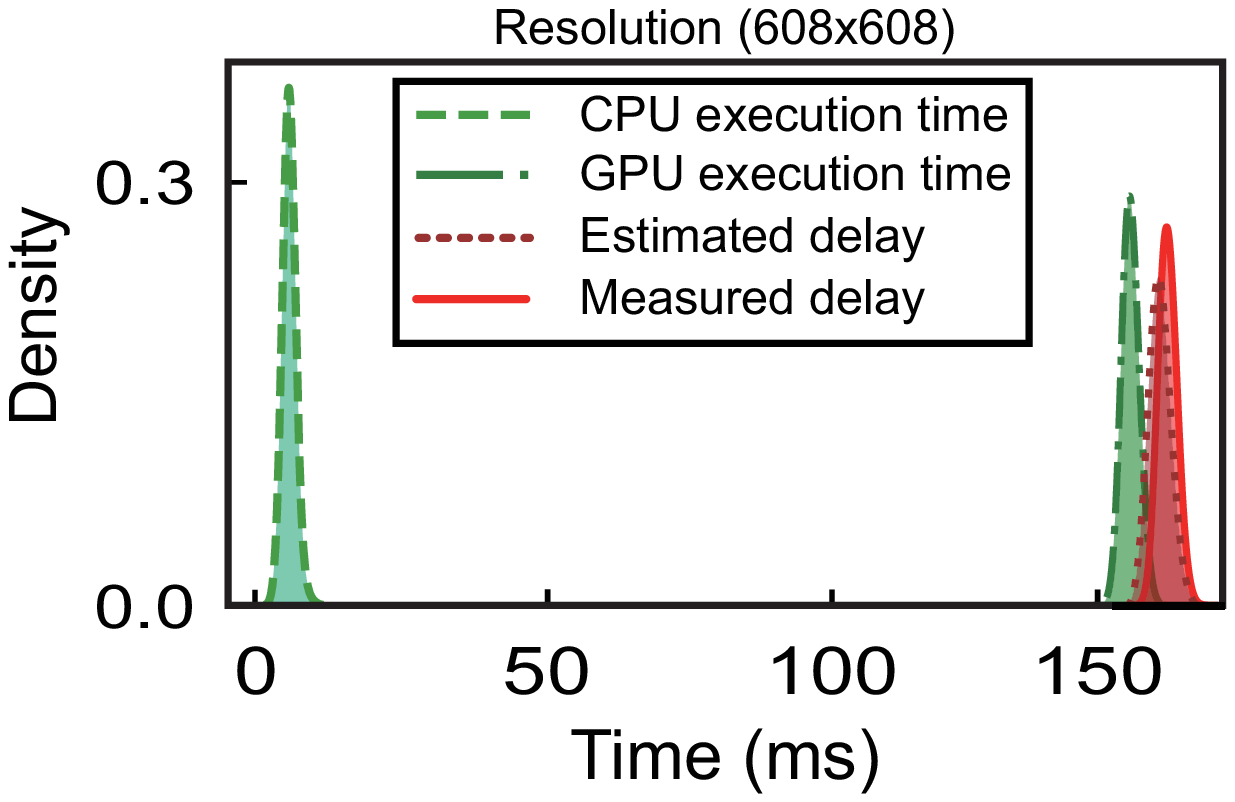}}
\\
\subfloat[$\mathcal{D}_{disp}(n_{obj})$]{\label{fig:prof_d}\includegraphics[width=.23\textwidth]{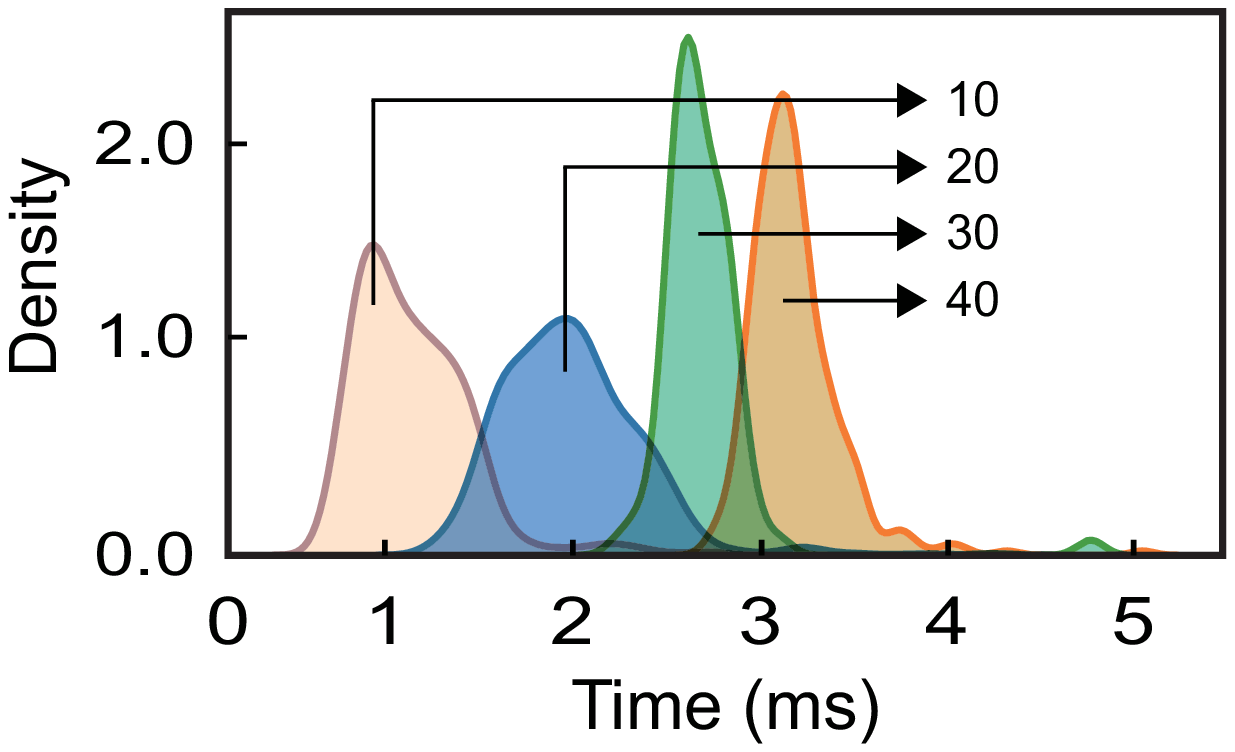}}
\quad
\subfloat[$e_{disp}$, $b_{disp}$, and $d_{disp}$]{\label{fig:prof_f4}\includegraphics[width=.23\textwidth]{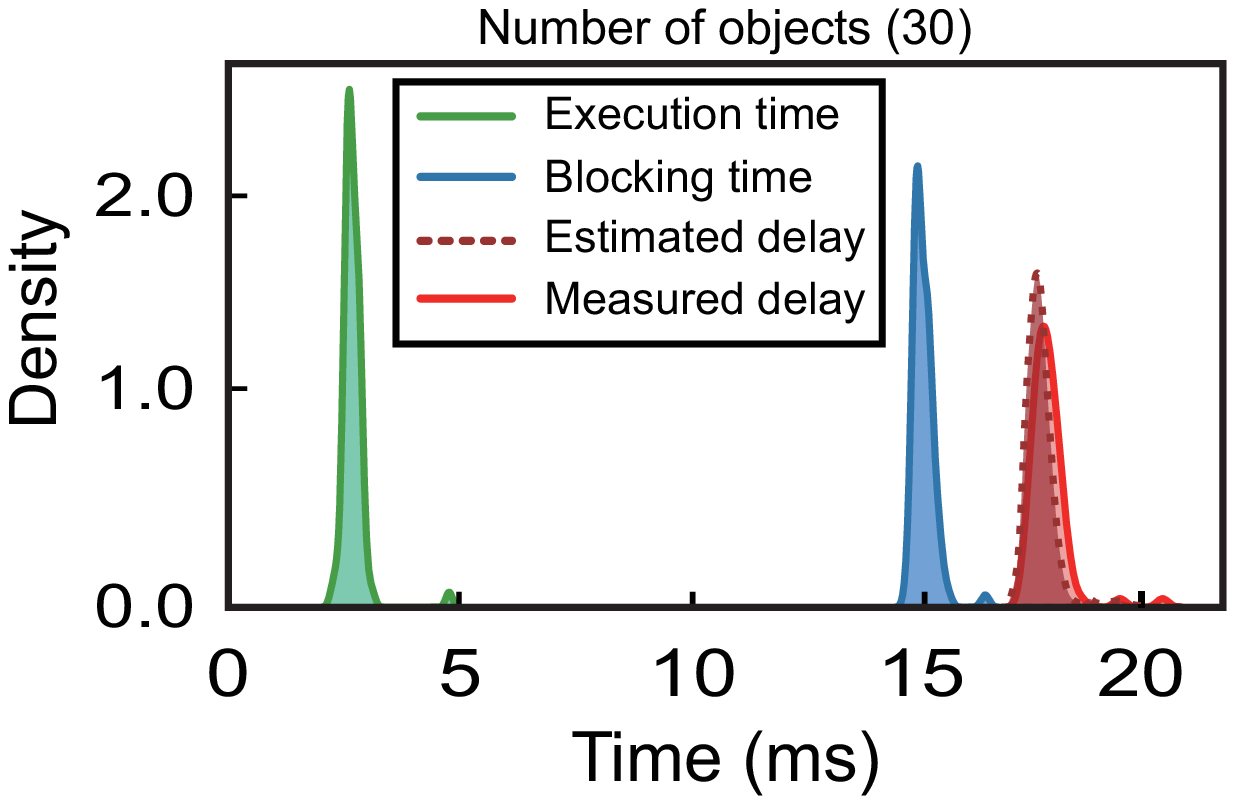}}
\caption{Profiling results from YOLO v3.}
\label{fig:profile}
\vspace{-0.5cm}
\end{figure}

{\bf (i) Fetch:} The fetch module is responsible for retrieving an image frame from the queue. However, because V4L2 supports the streaming I/O method where only pointers to buffers are exchanged between the driver and applications~\cite{v4l2}, there is no actual memory copy for data transfer. Instead, most of its execution time is spent in resizing an input image to match the resolution of a DNN input layer. Therefore, the execution time of the fetch module, $e_{fetch}$, can be approximated as follows:
\begin{equation}
    e_{fetch} \sim \mathcal{D}_{fetch}(r_{in}, r_{out}),
\end{equation}
where $\mathcal{D}_{fetch}(\cdot)$ is a probability distribution that is a function of input resolution ($r_{in}$) and output resolution ($r_{out}$) for resizing. Fig.~\ref{fig:prof_f} presents the profiling results of $\mathcal{D}_{fetch}(\cdot)$ with varying ($r_{in}$, $r_{out}$). Now, the delay of the fetch module, denoted by $d_{fetch}$, is defined as the time elapsed between its release (starting fetch thread) and its completion (ending image resize), which can be generalized as follows:
\begin{equation}
    d_{fetch} = e_{fetch} + b_{fetch},
\end{equation}
where $b_{fetch}$ is a blocking time factor. Note that, because no other background workload is assumed, no preemption delay is considered in $d_{fetch}$. Regarding $b_{fetch}$, a blocking system call in the fetch module may arise when an image is being retrieved from the queue. More specifically, when the fetch module tries to fetch an image while the queue happens to be empty, it is blocked until a new image arrives. Fig.~\ref{fig:prof_f2} shows the profiling results of $e_{fetch}$, $b_{fetch}$, and $d_{fetch}$ with fixed input and output resolutions presented above the plot. In the figure, $b_{fetch}$ = 0 at all times, meaning that there were no empty queue cases. However, this depends on the system status, in particular, the queue status, which will be discussed in Section~\ref{sec:queue_delay}.

{\bf (ii) Inference:} The inference module performs forward propagation in a neural network, which is decomposed into CPU and GPU parts. In the CPU part, it simply calls asynchronous GPU operations with CUDA function calls, which include data copies between CPU and GPU memory as well as GPU kernel functions for the neural network. Then the GPU operations are queued into a CUDA stream, which is a FIFO (First In, First Out) queue maintained by the CUDA runtime. The CPU thread blocks until the asynchronous GPU operations are completed by explicitly calling a CUDA synchronization function. Then the last GPU operation copies the detection results to the CPU memory. To run a single forward propagation, for example, YOLO v3 requires four copy operations and 322 GPU kernel operations (including repetitions). Note that the execution time of the inference module depends on the resolution of the input to the neural network $r_{nn}$, which is indicated by the profiling results in Fig.~\ref{fig:prof_i}. Thus, the CPU execution time of the inference module $e_{infer}^{CPU}$ and its GPU execution time $e_{infer}^{GPU}$ are defined by random variables as follows:
\begin{equation}
\begin{centering}
\begin{aligned}
    &e_{infer}^{CPU} \sim \mathcal{D}_{infer}^{CPU}(r_{nn})\textrm{ and}\\
    &e_{infer}^{GPU} \sim \mathcal{D}_{infer}^{GPU}(r_{nn}),
\end{aligned}
\end{centering}
\end{equation}
where $\mathcal{D}_{infer}^{CPU}(\cdot)$ and $\mathcal{D}_{infer}^{GPU}(\cdot)$ are probability distributions for CPU and GPU execution times, respectively, both of which are dependent on $r_{nn}$. For notational simplicity, the (unified) execution time of the inference module, $e_{infer}$, is defined as $e_{infer} = e_{infer}^{CPU} + e_{infer}^{GPU}$. Because there is no additional blocking factor, the delay of the inference module, $d_{infer}$, is given by
\begin{equation}
    d_{infer} = e_{infer} \sim \mathcal{D}_{infer}(r_{nn}),
\end{equation}
where $\mathcal{D}_{infer}(\cdot)$ can be estimated by the convolution of $\mathcal{D}_{infer}^{CPU}(\cdot)$ and $\mathcal{D}_{infer}^{GPU}(\cdot)$, as follows:
\begin{equation}
    \mathcal{D}_{infer}(\cdot) = \mathcal{D}_{infer}^{CPU}(\cdot) \otimes \mathcal{D}_{infer}^{GPU}(\cdot).
    \label{eq:dinfer}
\end{equation}
Fig.~\ref{fig:prof_f3} shows the profiling results with a fixed $r_{nn}$, where the estimated delay by Eq.~\eqref{eq:dinfer} closely matches the measured delay. As shown in the figure, most of the execution time is spent in the GPU part.

{\bf (iii) Display:} The display module visualizes the detection results by showing images with bounding boxes around each detected object. To accomplish this, the display module takes the following steps: first, the raw detection results with various confidence levels are filtered by a given confidence threshold. Next, irrelevant duplicates are joined (or removed) by the non-maximal suppression algorithm. Further, this module draws rectangles around each detected object. Thus, the execution time of the display stage, denoted by $e_{disp}$, is dependent on the number of detected objects $n_{obj}$, and is defined as
\begin{equation}
    e_{disp} \sim \mathcal{D}_{disp}(n_{obj}).
\end{equation}
In our measurements, we found that the number of raw detections and the final number of detected objects have a highly positive correlation. Thus, instead of using two variables, we use only the final number of detected objects when defining the probability distribution. Fig.~\ref{fig:prof_d} depicts the execution time distributions of the display stage against different values of $n_{obj}$. In addition, a blocking factor $b_{disp}$ exists, which is caused by calling the OpenCV function that renders images on a display. Thus, the delay of the display module, $d_{disp}$, can be estimated by
\begin{equation}
    d_{disp} = e_{disp} + b_{disp}.
\end{equation}
In the profiling results in Fig.~\ref{fig:prof_f4}, $b_{disp}$ is significant because it involves the GUI subsystem of the operating system (e.g., X-window), which processes related pending events on the display window. The figure specifically shows the profiling results for $n_{obj}$ = 30. For the worst-case analysis, we assume $n_{obj}$ = 30 as the maximum number of detectable objects. Note that Darknet has a configurable limit in its configuration file. Moreover, even commercial object detection systems impose a limit on the possible number of detectable objects in their specifications. Note that although the display stage above explained  is not always necessary in production vehicles, we consider it as a representative example of indispensable post-processing works.

\begin{figure}
\psfrag{s}[0.3]{$s$}
\psfrag{d}[0.3]{$d_{detector}$}
\centerline{\includegraphics[scale=0.54]{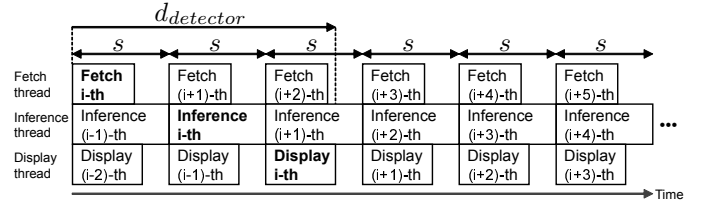}}
\caption{Multithreaded pipeline architecture.}
\label{fig:pipeline}
\vspace{-0.5cm}
\end{figure}

{\bf Multithreaded pipeline architecture.} In Darknet, the above three modules are organized with a multithreaded pipeline architecture to exploit multicore processors, using the fork-join model~\cite{conway1963multiprocessor}. There are three parallel threads, one for each module: (i) fetch thread, (ii) inference thread, and (iii) display thread. At the beginning of an object detection cycle, the three threads are forked and then joined at the end of the loop to begin the next cycle. Fig.~\ref{fig:pipeline} shows the timeline of the multithreaded pipeline architecture. In general, during the $i$-th object detection cycle, the $i$-th image frame is fetched while the ($i$-1)-th image frame is processed by the neural network, and the detection result of the ($i$-2)-th image is displayed on the screen. In the figure, since the inference is the longest, the fetch and display threads should wait for the inference thread to complete at the end of each object detection cycle. Thus, the object detection {\em service interval}, $s$, is now generalized as
\begin{equation}
    s = \textrm{max}(\{d_{fetch}, d_{infer}, d_{disp}\}),
    \label{eq:smax}
\end{equation}
which is also referred to as object detection {\em cycle time} depending on the context. Regarding $s$, according to the profiling results in Fig.~\ref{fig:profile}, $d_{infer}$ is significantly longer than $d_{fetch}$ and $d_{disp}$. Thus, in our experimental platform, we can simply assume a service interval distribution $\mathcal{S}$ as
\begin{equation}
    \mathcal{S} = \mathcal{D}_{infer}(\cdot).
\end{equation}
However, in general, the service interval distribution $\mathcal{S}$ of a detector subsystem by Eq.~\eqref{eq:smax} relies on possible combinations of the values of three random variables $\{d_{fetch}, d_{infer}, d_{disp}\}$; the distribution can be defined as the set of the maximum among these three values for each of all possible combinations. Assuming $d_{fetch}=A$, $d_{infer}=B$, and $d_{disp}=C$, the corresponding probability is the product of the three respective probabilities, i.e., $\textrm{Pr}\left\{d_{fetch} = A\right\}\times\textrm{Pr}\left\{d_{infer} = B\right\}\times\textrm{Pr}\left\{d_{disp} = C\right\}$. The final $\mathcal{S}$ can be obtained by summing all the probabilities for a given service interval. Based on $\mathcal{S}$, the delay of the detector subsystem, denoted by $d_{detector}$, can be represented as follows:
\begin{equation}
    d_{detector} = 2s + d_{disp},
\end{equation}
signifying two object detection cycles plus one delay of the display thread, as depicted in Fig.~\ref{fig:pipeline}. Then the minimum and maximum detector subsystem delays can be calculated as
\begin{equation}
\begin{centering}
\begin{aligned}
&d_{detector}^{min} = 2s^{min} + d_{disp}^{min}\textrm{ and}\\
&d_{detector}^{max} = 2s^{max} + d_{disp}^{max}.
\end{aligned}
\end{centering}
\label{eq:d_detector}
\end{equation}

\subsection{Queue Delay}
\label{sec:queue_delay}
Based on the image arrival interval distribution $\mathcal{A}$ and the object detection service interval distribution $\mathcal{S}$ discussed in Section~\ref{sec:camera_delay} and Section~\ref{sec:detector}, we can analyze the queue delay $d_{queue}$, which is defined as the time each image frame spends inside the queue.
Darknet uses the OpenCV VideoCapture library for queue management with its default queue size $Q$=4. Recall that a full queue cannot accommodate additional incoming image frames. For the analysis of $d_{queue}$, we have to consider the following three cases:

{\bf Case 1: $\textrm{min}(\mathcal{A}) > \textrm{max}(\mathcal{S})$.} In this case, object detection service intervals are always shorter (faster) than image arrival intervals. Then, from the perspective of the fetch thread, every time it attempts to retrieve an image frame, the queue is empty, thereby the fetch thread is blocked until an image frame arrives. Hence,
\begin{equation}
d_{queue} = 0.
\label{eq:d_q_case1}
\end{equation}
In this case, $b_{fetch}$ is eventually lengthened until $d_{fetch} = b_{fetch} + e_{fetch} = a$, where $a$ is the image arrival interval. As a result, object detection cycle times are regulated by the image arrival intervals, regardless of $d_{infer}$ and $d_{disp}$.

{\bf Case 2: $\textrm{max}(\mathcal{A}) < \textrm{min}(\mathcal{S})$.} In this case, image arrival intervals are always shorter (faster) than object detection service intervals, such that whenever the fetch thread tries to retrieve an image frame, the queue is full. Thus, $b_{fetch}$=0 at all times. For the queue delay, after an image enters the tail of the queue, it requires $Q$ object detection cycles (i.e., fetches) until it is dequeued. Thus, it takes roughly $Q \cdot s^{max}$ in the worst case.
To be precise, the maximum queue delay is calculated as follows:
\begin{equation}
    d_{queue}^{max} = Q \cdot s^{max} - (d_{tran}^{min} - M \cdot U).
    \label{eq:d_queue_final}
\end{equation}
The subtracted part reflects the minimum time that elapses between the beginning of an object detection cycle (i.e., making room to take newly captured images) and the actual arrival of an image frame. Note that the maximum queuing decision jitter ($M \cdot U)$ should be considered because, at most, that amount of an image frame can be transmitted, after its capture, even before the object detection cycle begins. 
In the best case, the value of $d_{queue}^{min}$ is calculated as follows:
\begin{equation}
d_{queue}^{min} = Q \cdot s^{min} - (d_{tran}^{max} + C).
\label{eq:d_queue_final2}
\end{equation}
This happens when an object appears just past the capture instant and has been successfully taken at the next capture instant. Then the delay of up to $C$ is accounted for in the capture delay, instead of the queue delay. In this case, we assume a zero queuing decision jitter to calculate the minimum queue delay.

{\bf Case 3: all other conditions.} In this case, the queue status is non-deterministic. Its long-term average can be estimated only when the averages of the two distributions are equal. Moreover, because the queue length is limited within [0, $Q$], the queue status at a certain instant cannot be estimated. Thus, the queue delay can be either $d_{queue}^{min}$= 0, as in {\bf Case 1}, in the best case, or can be as much as Eq.~\eqref{eq:d_queue_final}, as in {\bf Case 2}, in the worst case.

\section{End-to-End Delay Optimization}
\label{sec:optimization}
Based on the analysis thus far, this section explains our approach to minimize the end-to-end delay in our object detection system. More specifically, the following three techniques will be explained: (i) on-demand capture, (ii) zero-slack pipeline, and (iii) contention-free pipeline. Fig.~\ref{fig:opt} shows how these techniques incrementally reduce the end-to-end delay from the original (vanilla) architecture to our final architecture. Note that the effect of our optimization methods depends on the relative execution times of the three threads. For ease of explanation, we assume that $d_{infer}$ is significantly longer than $d_{fetch}$ and $d_{disp}$, as illustrated in Fig.~\ref{fig:arch_vanilla}. In this case, the object detection cycle time $s$ is {\em dominated} by $d_{infer}$, which is the most common case in embedded systems. The general applicability of our approach will be discussed in Section~\ref{sec:general}.

\subsection{On-demand Capture}
\label{sec:on-demand}
From Eqs.~\eqref{eq:d_queue_final} and \eqref{eq:d_queue_final2} in Section~\ref{sec:queue_delay}, it is apparent that we can reduce the queue delay by decreasing the queue size $Q$, especially for {\bf Case 2}, and partly for {\bf Case 3}. Note that even for {\bf Case 1}, reducing $Q$ has no negative side effect because it already works as though no queue exists. Motivated by the above observation, we can completely remove the queue, such that we only receive image frames when they are needed. We call this the {\em on-demand capture} method. For the implementation, we modified the OpenCV VideoCapture library. Listing~\ref{lst:with_queue} shows the pseudo code of the original OpenCV implementation where, in the initialization phase, a fixed queue size $Q$ is requested from the driver with the ioctl command (\texttt{REQBUFS}) (Line 1). Then $Q$ empty buffers, i.e., image holders, are put into the queue using the ioctl command (\texttt{QBUF}) (Line 2-3). After that, the fetch loop begins. Initially, it checks for the existence of an available image frame by making a select system call (Line 6). Upon returning from the select system call, it dequeues a buffer containing an image using the ioctl command (\texttt{DQBUF}) (Line 7), and then queues an empty buffer using the ioctl command (\texttt{QBUF}) (Line 8) again. It begins the next loop instantly. For comparison, Listing~\ref{lst:odf} shows our modified implementation with $Q$=1 (Line 1), which inserts an empty buffer into the queue using the ioctl command (\texttt{QBUF}) (Line 6) just before making the select system call. After the select system call returns, notifying an image arrival (Line 7), the buffer with the image frame is dequeued by the ioctl command (\texttt{DQBUF}) (Line 8), and the next loop begins.

\vspace{0.7\baselineskip}
\noindent\begin{minipage}{.485\linewidth}
\begin{lstlisting}[language=C++, escapechar=@, caption={Original.}, label={lst:with_queue}, frame=tbrl]
 @\textbf{1}@  @\textbf{ioctl}@(fd, @\color{red}{\textbf{REQBUFS}}@, @\color{red}{\textbf{Q}}@);
 @\textbf{2}@  for(i=0; i<@\color{red}{\textbf{Q}}@; i++)
 @\textbf{3}@    @\textbf{ioctl}@(fd, @\color{red}{\textbf{QBUF}}@, buf);
 @\textbf{4}@  @\textbf{/* Image fetch loop */}@
 @\textbf{5}@  while(1){
 @\textbf{6}@    @\color{red}{\textbf{select}}@(fd);
 @\textbf{7}@    @\textbf{ioctl}@(fd, @\color{red}{\textbf{DQBUF}}@, buf);
 @\textbf{8}@    @\textbf{ioctl}@(fd, @\color{red}{\textbf{QBUF}}@, buf);
 @\textbf{9}@  }
\end{lstlisting}
\end{minipage}\hfill
\begin{minipage}{.485\linewidth}
\begin{lstlisting}[language=C++, escapechar=@, caption={On-demand capture.}, label={lst:odf}, frame=tbrl]
 @\textbf{1}@  @\textbf{ioctl}@(fd, @\color{red}{\textbf{REQBUFS}}@, @\color{red}{\textbf{1}}@);
 @\textbf{2}@
 @\textbf{3}@
 @\textbf{4}@  @\textbf{/* Image fetch loop */}@
 @\textbf{5}@  while(1){
 @\textbf{6}@    @\textbf{ioctl}@(fd, @\color{red}{\textbf{QBUF}}@, buf);
 @\textbf{7}@    @\color{red}{\textbf{select}}@(fd);
 @\textbf{8}@    @\textbf{ioctl}@(fd, @\color{red}{\textbf{DQBUF}}@, buf);
 @\textbf{9}@  }
\end{lstlisting}
\end{minipage}

\begin{figure}
\centering
\subfloat[Vanilla (original) architecture]{\label{fig:arch_vanilla}\includegraphics[width=.46\textwidth]{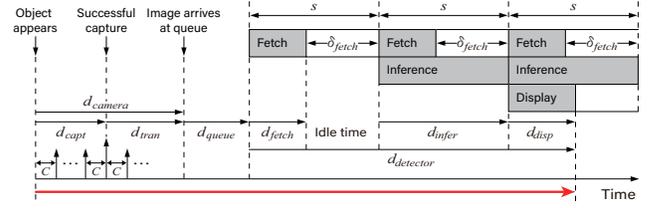}}\\
\subfloat[$+$ On-demand capture: queue delays are removed.]{\label{fig:arch_on_demand}\includegraphics[width=.46\textwidth]{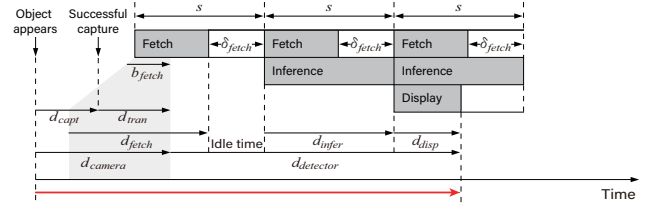}}\\
\subfloat[$+$ Zero-slack pipeline: idle time gaps are removed.]{\label{fig:arch_zero_slack}\includegraphics[width=.46\textwidth]{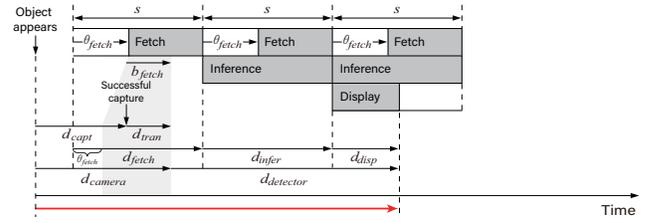}}\\
\subfloat[$+$ Contention-free pipeline: memory contention is removed.]{\label{fig:arch_contention}\includegraphics[width=.46\textwidth]{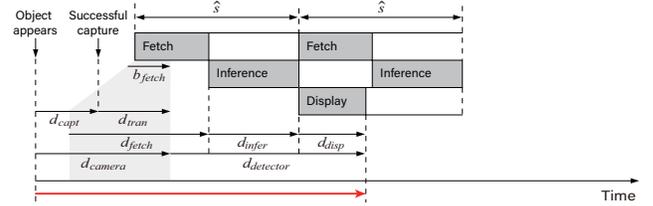}}\\
\caption{End-to-end delay optimization methods. The thick red arrows below the time axes depict the end-to-end delays.}
\label{fig:opt}
\vspace{-0.5cm}
\end{figure}

Fig.~\ref{fig:arch_vanilla} illustrates the vanilla architecture with a queue, whereas Fig.~\ref{fig:arch_on_demand} depicts our on-demand capture architecture. By comparing the two, it can be seen that our on-demand capture architecture no longer has a queue delay. Instead, it experiences a blocking time $b_{fetch}$ when it makes the select system call, as in Listing~\ref{lst:odf}. More specifically, the minimum and maximum blocking times are determined as follows:
\begin{equation}
\begin{centering}
\begin{aligned}
&b_{fetch}^{min} = d_{tran}^{min} - M \cdot U\textrm{ and}\\
&b_{fetch}^{max} = d_{tran}^{max} + C.
\end{aligned}
\end{centering}
\end{equation}
Note that they are the same as the subtracted parts of Eq.~\eqref{eq:d_queue_final} and Eq.~\eqref{eq:d_queue_final2}, respectively, which correspond to waiting times from the beginning of an object detection cycle until a new image frame, which will be used later, arrives at the queue, even though there is no blocking. In contrast, in the on-demand capture architecture, the fetch thread is blocked for the exact same amount of time as the waiting time, and once unblocked, the arriving image frame is fed directly into the fetch thread with no queue delay.

In Fig.~\ref{fig:arch_on_demand}, by the on-demand capture method described above, the queue delay is completely removed as
\begin{equation}
    d_{queue} = 0.
\end{equation}
As a side effect, in the figure, note that $d_{camera}$ and $d_{fetch}$ are somewhat overlapped by $b_{fetch}$, because the fetch thread is blocked waiting for an image arrival for $b_{fetch}$ at its beginning. Thus, when calculating the detector subsystem delay, the overlapped amount of time ($b_{fetch}$), which is already accounted for in $d_{camera}$, should be subtracted from the original delay equations in Eq.~\eqref{eq:d_detector}, so we have the following update to the detector subsystem delay:
\begin{equation}
\begin{centering}
\begin{aligned}
    &d_{detector}^{min} = 2s^{min} + d_{disp}^{min} - b_{fetch}^{max}\textrm{ and}\\
    &d_{detector}^{max} = 2s^{max} + d_{disp}^{max} - b_{fetch}^{min}.
\end{aligned}
\end{centering}
\label{eq:d_detector_final}
\end{equation}

\subsection{Zero-slack Pipeline}
\label{sec:zero-slack}
From Fig.~\ref{fig:arch_on_demand}, we can observe significant idle time gaps between the end of fetches and the beginning of the next object detection cycles, caused by the different lengths of the three pipeline stages, i.e., $d_{fetch}$, $d_{infer}$, and $d_{disp}$. If we can remove these idle times, the end-to-end delay can be further reduced. To that end, we introduce the notions of the slack and offset of the fetch thread, as follows:
\begin{itemize}
    \item {\bf Fetch-thread slack ($\delta_{fetch}$):} time gaps between the end of a fetch thread and the beginning of the next object detection cycle.
    \item {\bf Fetch-thread offset ($\theta_{fetch}$):} an artificial constant delay at every release of the fetch thread.
\end{itemize}
By accounting for an offset $\theta_{fetch}$, $d_{fetch}$ is increased to
\begin{equation}
    d_{fetch} = \theta_{fetch} + e_{fetch} + b_{fetch}.
\end{equation}
Then its slack can be calculated by
\begin{equation}
    \delta_{fetch} = s - d_{fetch},
\end{equation}
where $s$ denotes the object detection cycle time. Thus, we can increase $\theta_{fetch}$ until $\delta_{fetch}$ reaches zero (zero-slack), which eventually eliminates idle time gaps, as illustrated in Fig.~\ref{fig:arch_zero_slack}. Then the detector subsystem delay is further reduced from Eq.~\eqref{eq:d_detector_final} to
\begin{equation}
\begin{centering}
\begin{aligned}
    &d_{detector}^{min} = 2s^{min} + d_{disp}^{min} - b_{fetch}^{max} - \theta_{fetch}\textrm{ and}\\
    &d_{detector}^{max} = 2s^{max} + d_{disp}^{max} - b_{fetch}^{min} - \theta_{fetch}.
\end{aligned}
\end{centering}
\end{equation}
However, applying too much $\theta_{fetch}$ may increase the object detection cycle time $s$ itself. Thus, we must find an optimal $\theta_{fetch}$ that minimizes the end-to-end delay as much as possible while not increasing $s$. For that, $\theta_{fetch}$ is conservatively decided as
\begin{equation}
    \theta_{fetch} = s^{min} - e_{fetch}^{max} - b_{fetch}^{max},
\end{equation}
such that the increased $d_{fetch}$ barely touches the minimum object detection cycle time $s^{min}$ (i.e., $\textrm{max}(d_{fetch}) \approx s^{min}$). Due to the conservative decision, we cannot completely remove all idle time gaps. However, it is apparent that zero-slack pipeline with an appropriate $\theta_{fetch}$ reduces the end-to-end delay to some extent.

\subsection{Contention-free Pipeline}
\label{sec:contention-freeb}
\begin{figure}
\centering
\subfloat[Fetch]{\label{fig:con_fetch}\includegraphics[width=.16\textwidth]{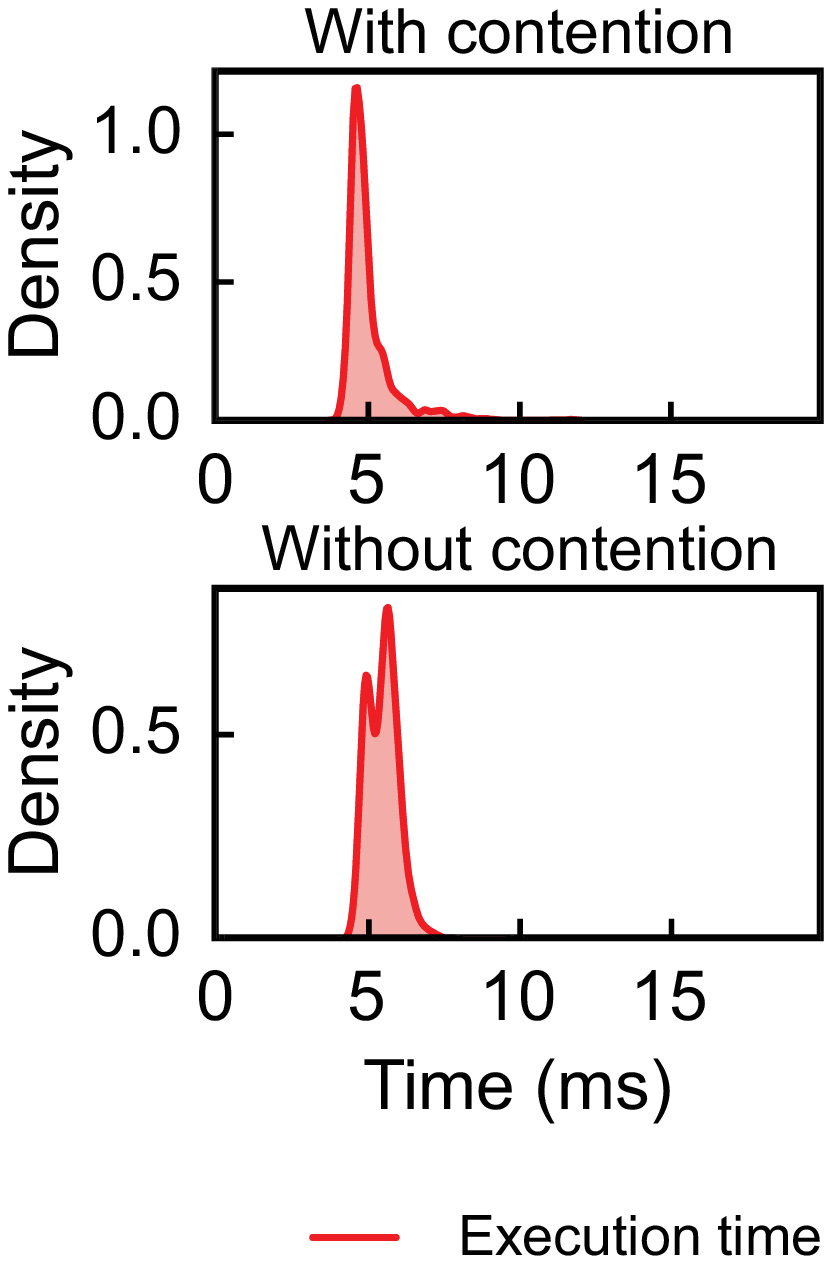}}
\subfloat[Inference]{\label{fig:con_inference}\includegraphics[width=.16\textwidth]{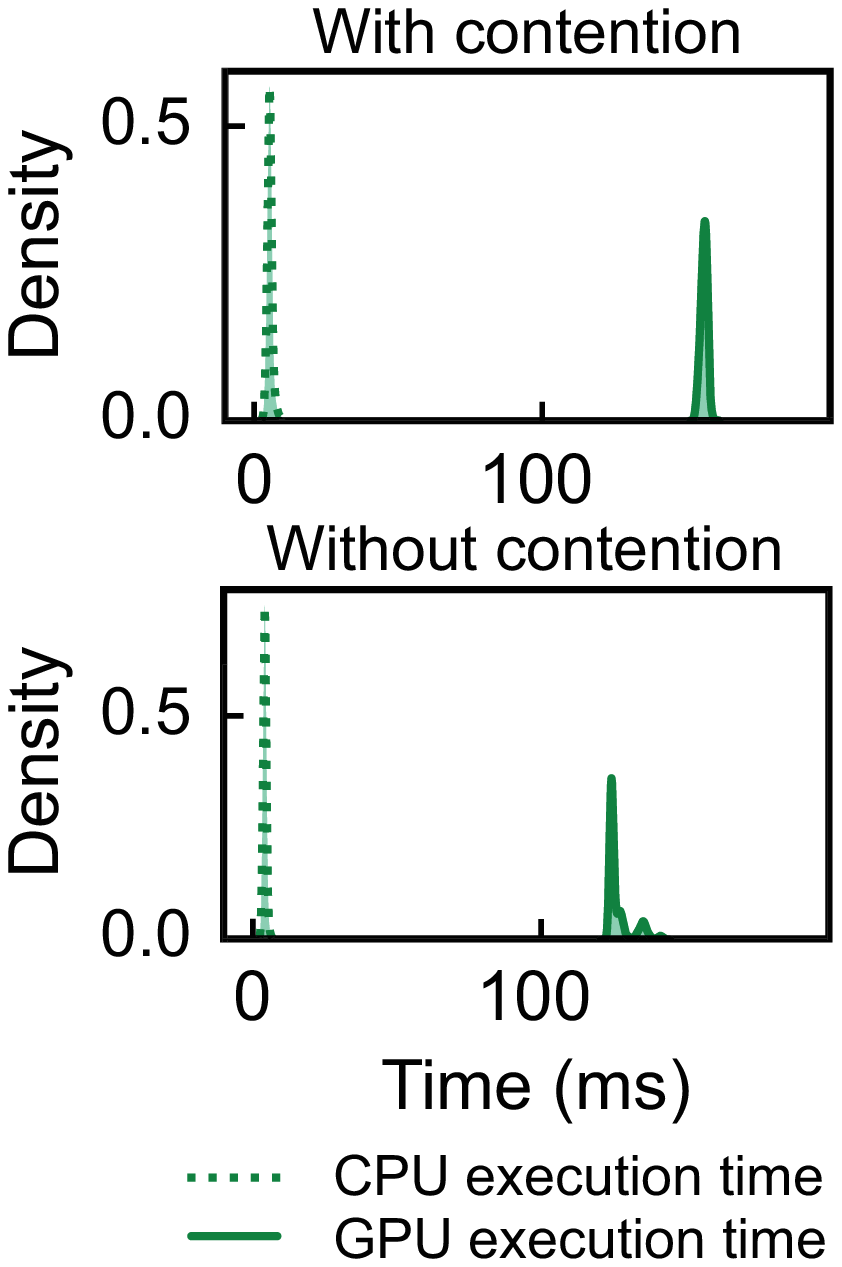}}
\subfloat[Display]{\label{fig:con_display}\includegraphics[width=.16\textwidth]{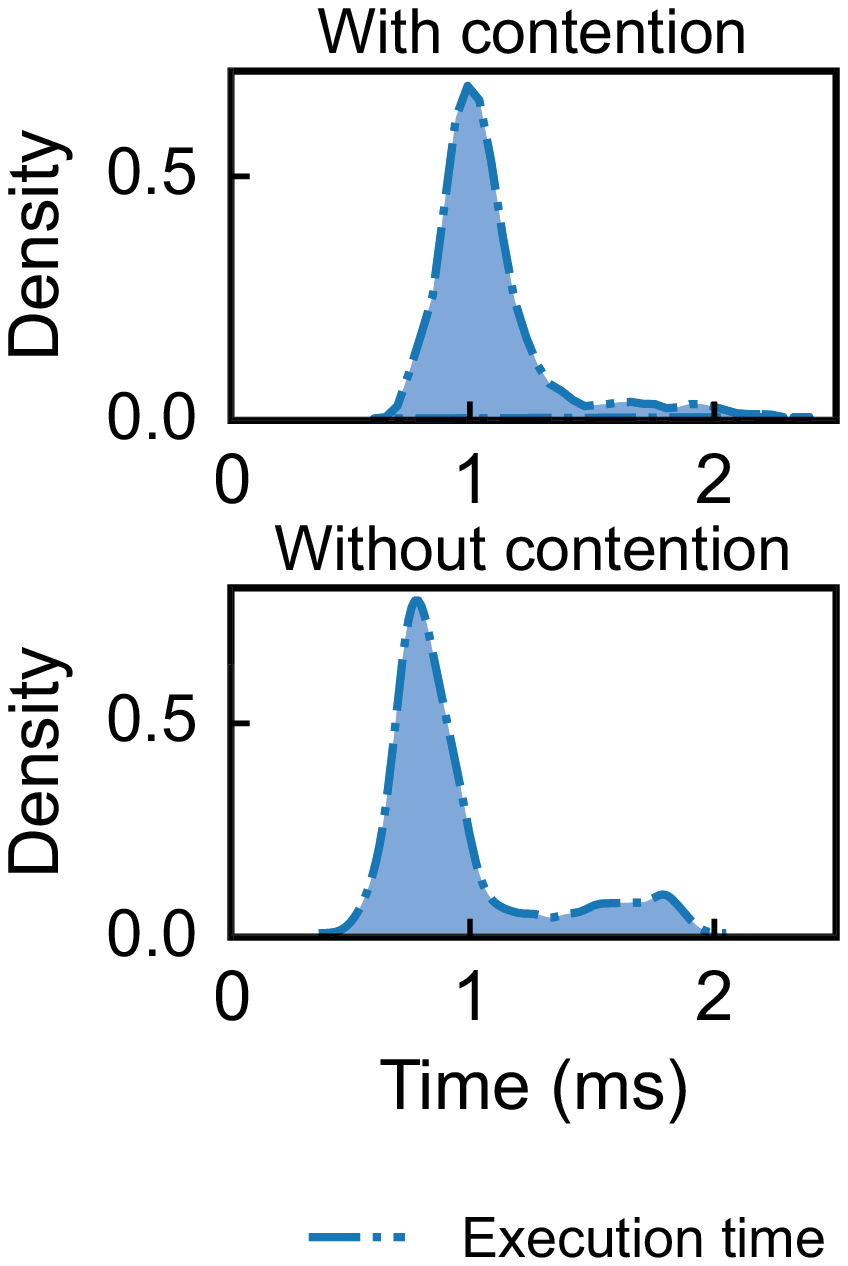}}
\caption{Comparison of execution times with and without shared memory bandwidth contention.}
\label{fig:contention}
\vspace{-0.5cm}
\end{figure}

When multiple threads execute on a shared memory multicore processor, there is significant memory bandwidth contention between concurrent threads~\cite{yun2013memguard, ali2019rt}. Moreover, when CPU threads run with GPU kernels on an integrated GPU, the effect is even worse~\cite{ali2017protecting}. Because Darknet uses a multithreaded pipeline architecture, our method is not free from this. To quantify the memory contention effect, Fig.~\ref{fig:contention} shows the execution time distributions with and without contention. The execution times with contention are measured in the vanilla architecture, whereas the execution times without contention are measured while sequentially executing the object detection stages. In particular, Fig.~\ref{fig:con_inference} shows a significant increase in execution time (about 28~ms) for the GPU executions on the inference thread. In contrast, the fetch and display threads in Figs.~\ref{fig:con_fetch} and \ref{fig:con_display} show relatively minor effects. Thus, it is clear that the inference thread suffers the most from the memory bandwidth contention. Based on this observation, our design choice is to isolate the inference thread to minimize the contention, whereas the fetch and display threads run concurrently as in the original architecture. Fig.~\ref{fig:arch_contention} shows the resulting contention-free pipeline architecture.

Unlike other proposed methods described so far, which reduce the end-to-end delay without any trade-off, the contention-free pipeline method is not free. As shown in Fig.~\ref{fig:arch_contention}, the object detection cycle time is changed (and possibly increased) to $\hat{s}$, which is defined as follows:
\begin{equation}
    \hat{s} = \textrm{max}(\{d_{fetch}, d_{disp}\}) + d_{infer}.
\label{eq:hats}
\end{equation}
However, because $d_{infer}$ is also decreased by contention-free executions, the increase from $s$ to $\hat{s}$ is marginal, which will be discussed in the experiments section. Then the detector subsystem delay is finally reduced as follows:
\begin{equation}
\begin{centering}
\begin{aligned}
&d_{detector}^{min} = \hat{s}^{min} + d_{disp}^{min} - b_{fetch}^{max}\textrm{ and}\\
&d_{detector}^{max} = \hat{s}^{max} + d_{disp}^{max} - b_{fetch}^{min}.
\end{aligned}
\end{centering}
\end{equation}


\subsection{Considering Other Real-Time Workloads}
\label{sec:rt}
Thus far, we have assumed no other workload in the system. This section discusses how our final pipeline architecture can be integrated with other real-time workloads. The most straightforward approach to ensure this integration is to isolate CPU cores required to parallelize each pipeline stage and a GPU, such that they are dedicated to our object detection system while the remaining CPU cores are shared by other workloads. However, this approach may still cause the same memory contention problem. To address this issue, we can use the RT-Gang scheduling framework~\cite{ali2019rt}, which eliminates inter-task interference by its one-gang-at-a-time policy. Here, a gang is a pre-defined group of tasks that is allowed to run in parallel. For example, in Fig.~\ref{fig:arch_contention}, we can define two gangs: (i) the fetch and display gang and (ii) the inference gang. Other real-time workloads can be grouped into a number of extra gangs, where all the gangs are scheduled by uniprocessor scheduling algorithms based on the RT-Gang framework.

Also, recall that the object detection cycle time $\hat{s}$ in Eq.~\eqref{eq:hats} is not a design parameter but the one that can be derived; we, therefore, should turn it into a design parameter (i.e., the period of our gangs) to use it in real-time scheduling. Then, if all the gangs are schedulable, the given object detection period will be guaranteed. Regarding the end-to-end delay, the following constraints should be met to benefit from its minimization: (i) within a gang period, the fetch and display gang should be completed prior to the beginning of the inference gang; (ii) the display thread should not be preempted by other real-time workloads. We set the fetch and display gang as the task with the highest priority in the system so as to satisfy the above constraints; this modification results in minimal end-to-end delay. The actual implementation and evaluation with the RT-Gang framework is left as our future work.

\section{General Applicability of Our Methods}
\label{sec:general}
This section discusses how our approach can be applied in more general settings in the following aspects:

{\bf To other object detection frameworks.} Although we use Darknet throughout this study, our analysis and optimization methods regarding the camera subsystem and the queue delay can be applied to other object detection frameworks without modification. Further, because most other frameworks are designed without considering multicore processors, their detector subsystem can be enhanced by adapting our approach along with Darknet's multithreaded pipeline architecture. Thus, we consider that our solution can be generally used for the industry when building their production object detection systems, regardless of the framework they use.

{\bf To other DNNs.} This study uses the YOLO DNNs because we are targeting object detection systems. However, as we do not exploit any DNN-specific features of YOLO, our methods can even be applied to other perception tasks (e.g., lane detection) with different DNN architectures. Nevertheless, we do not claim that our final architecture, as shown in Fig.~\ref{fig:arch_contention}, is generally optimal across all DNNs. For example, let us replace the YOLO inference stage with a lightweight lane detection DNN, which makes the pipeline stages well balanced, i.e., $d_{fetch}\approx d_{infer}\approx d_{disp}$. Then we cannot expect a meaningful delay reduction with our zero-slack pipeline. Thus, each optimization technique should be selectively applied after investigating the execution patterns of the target DNN.

{\bf To other hardware platforms.} We conduct experiments on an Nvidia Jetson AGX Xavier, having an integrated GPU with significant memory contention between the CPU and the GPU. Although our analysis framework generally works for any GPU-based hardware platform, certain optimization techniques may not always provide as much benefit as in our experimental platform. For example, on other hardware platforms with a dedicated GPU with its isolated memory subsystem, our contention-free method will not provide satisfactory delay reductions as in our experimental platform. However, note that Nvidia Jetson AGX Xavier is a typical example of embedded systems for developing computer vision applications.


\section{Experiments}
\label{sec:experiments}
\begin{table}
  \caption{Default system configurations.}
  \label{tab:configs}
\centering
\begin{tabular}{ll}
\toprule
 {\bf Parameter}&  {\bf Value}\\ 
 \hline
  Camera frame rate ($F$)          & 30~fps\\
  Camera cycle time ($C$)          & 33.333$\cdots$~ms\\
  Camera resolution ($X$$\times$$Y$) & 640$\times$480\\
  Bits per pixel ($P$)             & 16 (YUYV pixel format)\\
  Queue size ($Q$)                 & 4\\
  YOLO neural network resolution  & 608$\times$608\\
\toprule
\end{tabular}
\vspace{-0.5cm}
\end{table}

\subsection{Implementation}
\label{sec:impl}

Our \underline{R}eal-\underline{T}ime \underline{O}bject \underline{D}etector (R-TOD\footnote{The source code is available at \url{https://github.com/aveeslab/r-tod}.}) is implemented on an Nvidia Jetson AGX Xavier with 16~GB RAM, an 8-core ARM CPU, and a 512-core integrated Volta GPU. We use the Logitech C930e USB camera in the USB 2.0 high-speed mode. Among many, we selected the 640$\times$480 resolution, 30~fps frame rate, and YUYV 16-bpp pixel format as our default camera configuration. As our software platform, we use Nvidia Ubuntu Linux-18.04 with JetPack-4.2.2, which is an SDK (Software Development Kit) for AI and computer vision applications, and a custom compiled OpenCV-3.3.1 library. We use the UVC camera driver in Linux. With regard to the application side, as the baseline object detector, the source code available at \url{https://github.com/AlexeyAB/darknet} is used. Note that this is the most well-known fork of the original YOLO repository. In recent times, it has been regarded as the official repository because the original author of YOLO stopped his research in computer vision. Furthermore, for neural networks, we use YOLO v2, v3, and v4 models pre-trained using the MS COCO dataset, which are available in the same GitHub repository. Finally, we use 608$\times$608 as the default input size of the neural network.

{\bf (i) On-demand capture.} The OpenCV VideoCapture library is modified to make the queue size a tunable parameter. Because there is no such API in the library, we added an environment variable \texttt{QLEN}. If \texttt{QLEN=0}, on-demand capture is activated.

{\bf (ii) Zero-slack pipeline.} We put a sleep function call at the beginning of the fetch thread. The amount of sleep $\theta_{fetch}$ is decided offline, and can be given by a command-line option when invoking the object detector.

{\bf (iii) Contention-free pipeline.} We no longer fork the three threads when beginning an object detection cycle. Instead, only the fetch and display threads are forked. After the two threads join, the inference thread is started alone in isolation. The next object detection cycle begins instantly after the completion of the inference thread.

\subsection{Evaluation of Our Analysis Methods}
\label{sec:eval_anal}
This section evaluates our end-to-end delay analysis framework presented in Section~\ref{sec:delay_analysis}. For the evaluation, YOLO v3 is used. However, note that we obtained similar evaluation results using YOLO v2 and v4 as well. Table~\ref{tab:configs} summarizes the default evaluation configuration, unless otherwise mentioned.

\begin{figure}
\begin{center}
\subfloat[$s$ = 100~ms]{\label{fig:d_cam_100}\includegraphics[width=.16\textwidth]{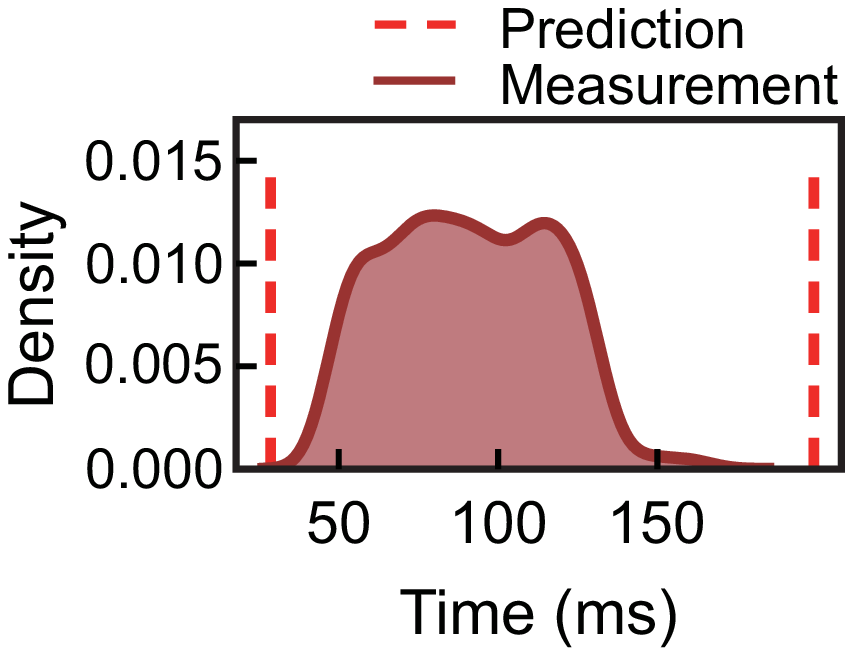}}
\subfloat[$s$ = 150~ms]{\label{fig:d_cam_150}\includegraphics[width=.16\textwidth]{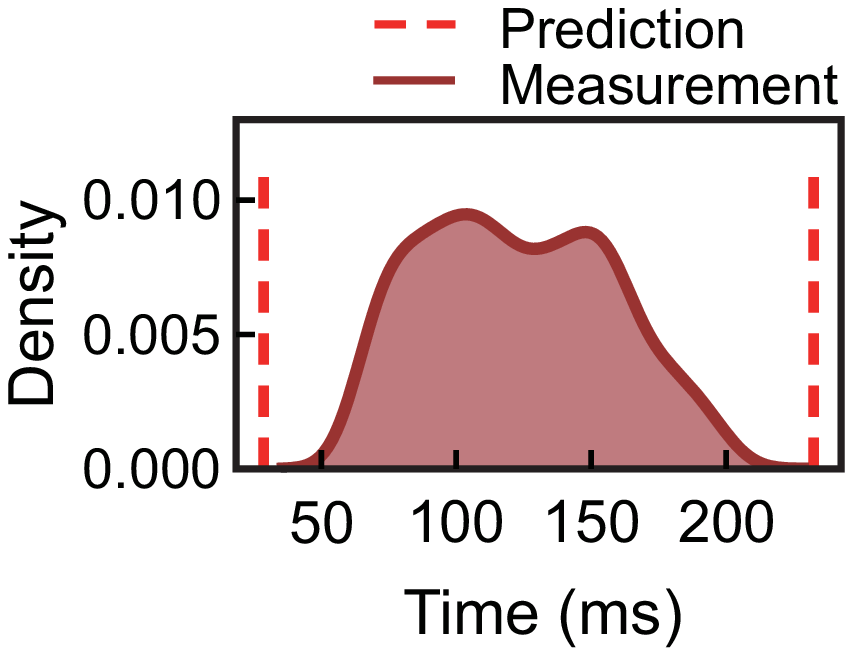}} 
\subfloat[$s$ = 200~ms]{\label{fig:d_cam_200}\includegraphics[width=.16\textwidth]{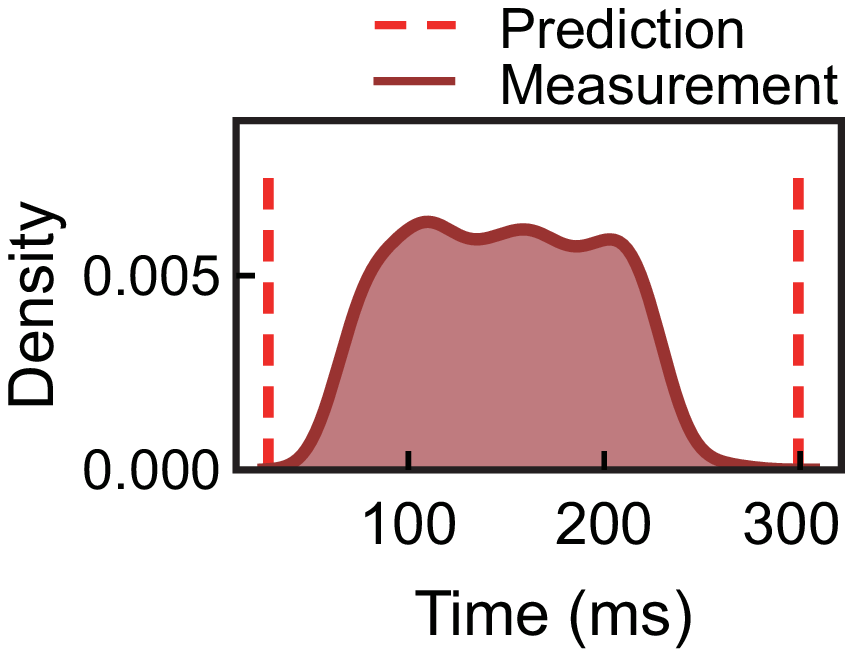}} 
\caption{Camera subsystem delays with varying service intervals (predictions and measurements).}
\label{fig:result_d_camera}
\end{center}
\vspace{-0.5cm}
\end{figure}

Fig.~\ref{fig:result_d_camera} shows the camera subsystem delay distributions where a camera subsystem delay is split into two parts: (i) $d_{capt}$ and (ii) $d_{tran}$. Between the two, $d_{tran}$ is actually measured, and $d_{capt}$ is randomly generated to simulate an object's random appearance between the previous and the current successful camera capture instants. Because the capture delay is tightly intertwined with the service interval of the detector subsystem, we measure the camera subsystem delays with varying service intervals. A dummy object detector with controllable artificial delays is used to control the service intervals. The figure shows that the camera subsystem delay distributions are within our predicted intervals as given by Eq.~\eqref{eq:d_camera_final}. Further, as reported in our analysis, the camera subsystem delay increases, from (a) to (c) in Fig.~\ref{fig:result_d_camera}, with longer object detection service intervals.

\begin{figure}
\begin{center}
\subfloat[\bf{Case 1}]{\label{fig:d_queue_case1}\includegraphics[width=.16\textwidth]{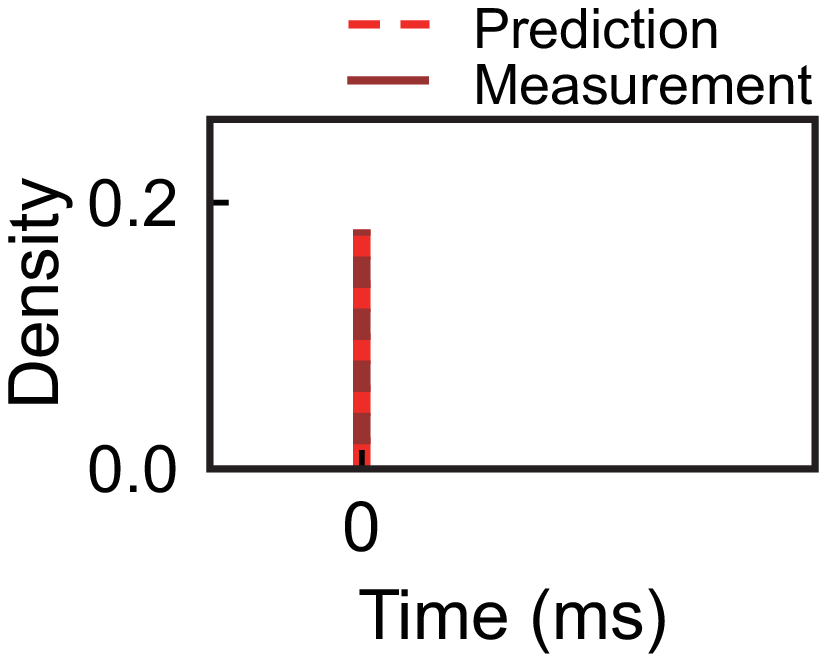}}
\subfloat[\bf{Case 2}]{\label{fig:d_queue_case2}\includegraphics[width=.16\textwidth]{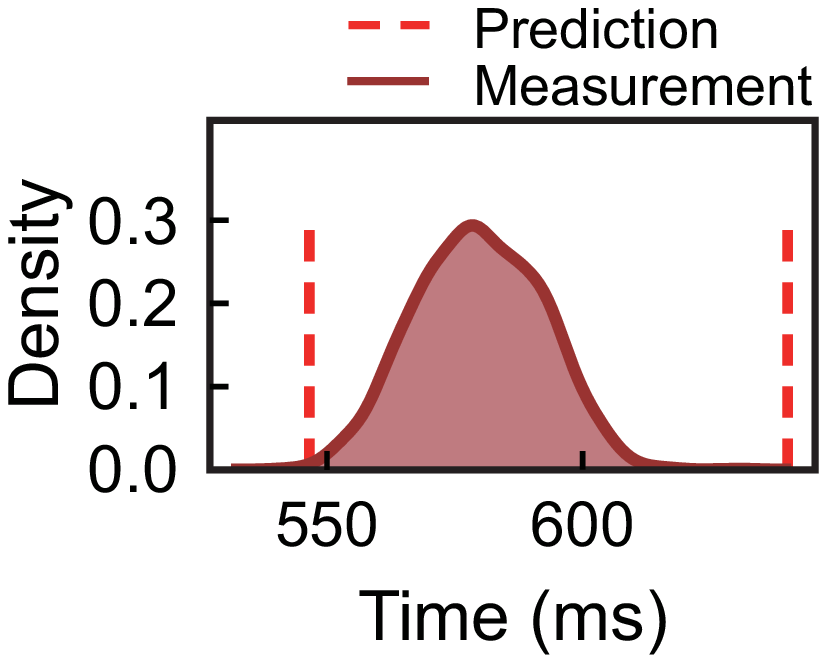}} 
\subfloat[\bf{Case 3}]{\label{fig:d_queue_case3}\includegraphics[width=.16\textwidth]{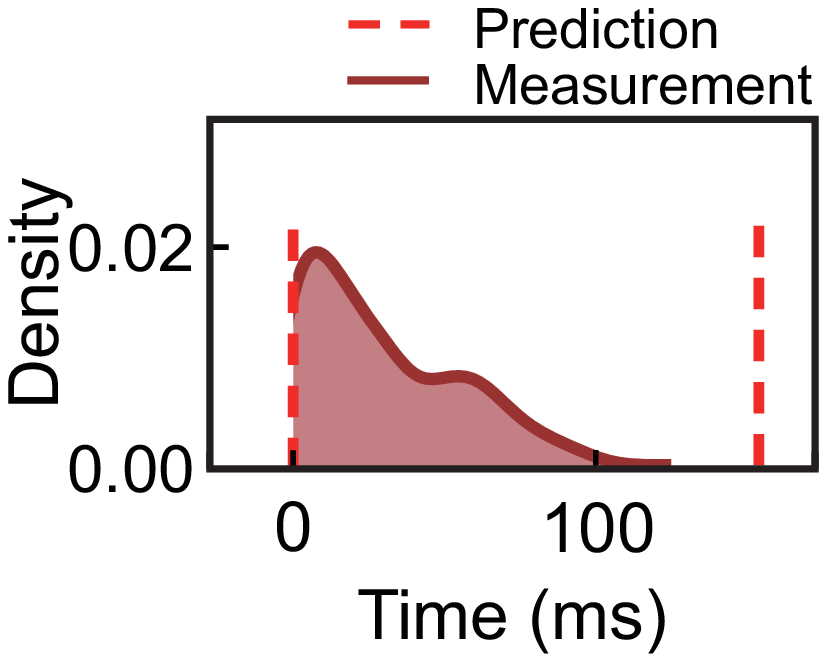}} 
\caption{Queue delays with the three cases (predictions and measurements).}
\vspace{-0.9cm}
\label{fig:result_d_queue}
\end{center}

\end{figure}

Fig.~\ref{fig:result_d_queue} shows the measurement results of the queue delays and our predictions for the three cases considered when analyzing the queue delay. In Fig.~\ref{fig:d_queue_case1} ({\bf Case 1}), $d_{queue}$ = 0 at all times as predicted in Eq.~\eqref{eq:d_q_case1}. In Fig.~\ref{fig:d_queue_case2} ({\bf Case 2}), the delay is predicted to be between 546~ms and 643~ms by Eq.~\eqref{eq:d_queue_final} and Eq.~\eqref{eq:d_queue_final2} that successfully estimate the actual measurements. In Fig.~\ref{fig:d_queue_case3} ({\bf Case 3}), the queue status is non-deterministic as reported in our analysis. The measurement results are irregularly distributed between 0~ms and 103~ms. Note that even for {\bf Case 3}, we can bound the minimum and maximum delays by reusing the analysis method of {\bf Case 1} in the best case and the analysis method of {\bf Case 2} in the worst case, respectively.

\begin{figure}
\begin{center}
\subfloat[v3 (544$\times$544)]{\label{fig:d_detector_544}\includegraphics[width=.16\textwidth]{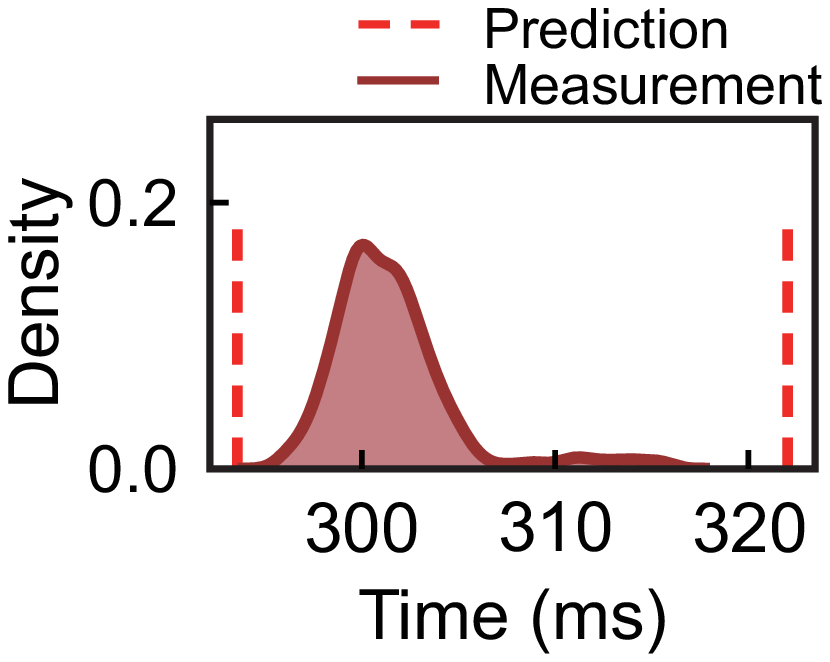}}
\subfloat[v3 (576$\times$576)]{\label{fig:d_detector_576}\includegraphics[width=.16\textwidth]{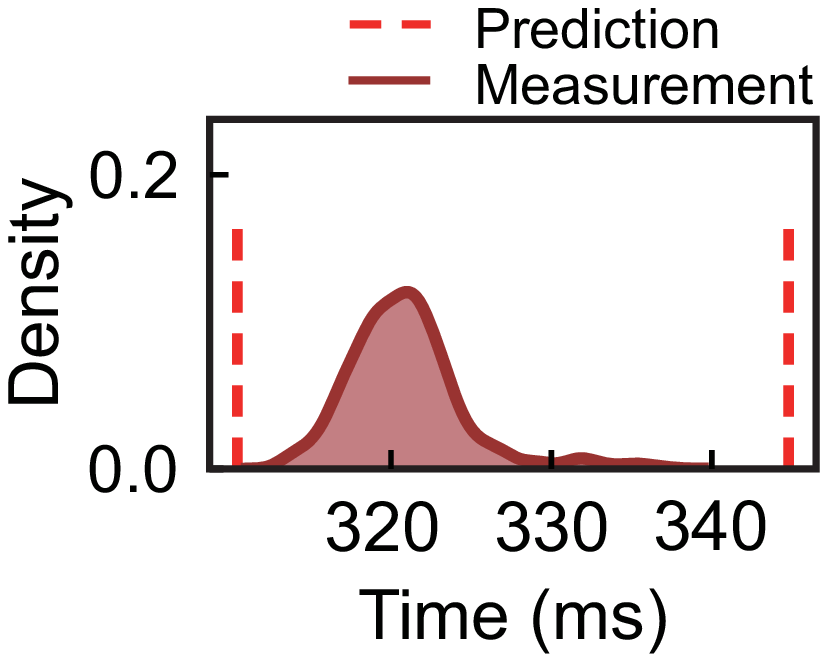}} 
\subfloat[v3 (608$\times$608)]{\label{fig:d_detector_608}\includegraphics[width=.16\textwidth]{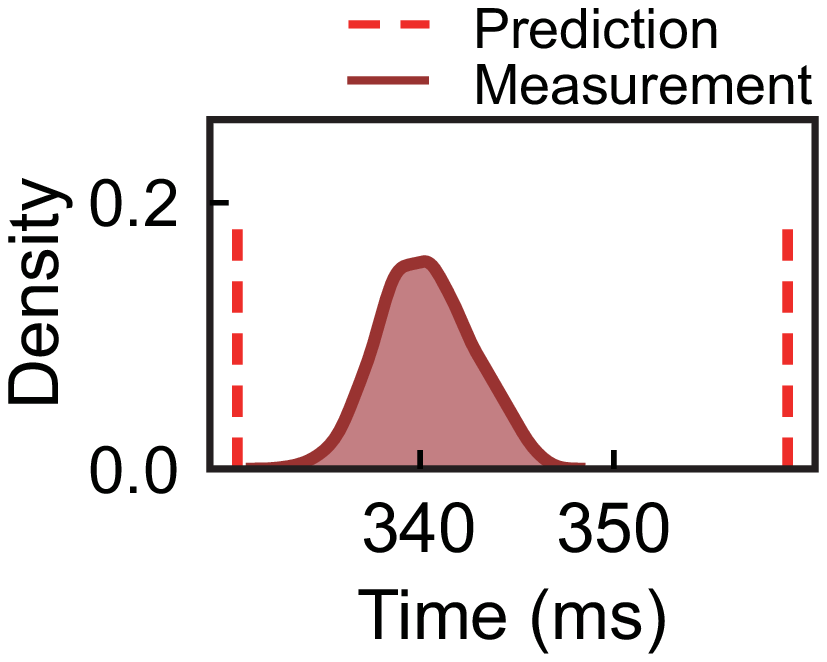}} 
\caption{Detector subsystem delays with varying neural network input resolutions (predictions and measurements).}
\label{fig:result_d_detector}
\vspace{-0.7cm}
\end{center}

\end{figure}

Fig.~\ref{fig:result_d_detector} presents the measurement results of the detector subsystem delay along with our predictions obtained using Eq.~\eqref{eq:d_detector}. The figures show that our predictions well estimate the detector subsystem delays with varying neural network resolutions. For the predictions, it is assumed that $n_{obj}$ = 30 for the worst-case analysis. During the measurement, however, the actual number of detected objects is mostly less than 15. This is one of the reasons for the unavoidable pessimism in the worst-case prediction.



\subsection{Evaluation of Our Optimization Methods}
\label{sec:eval_opt}

\begin{figure}
\begin{center}
\subfloat[End-to-end delays]{\label{fig:Qlen_delay}\includegraphics[width=.20\textwidth]{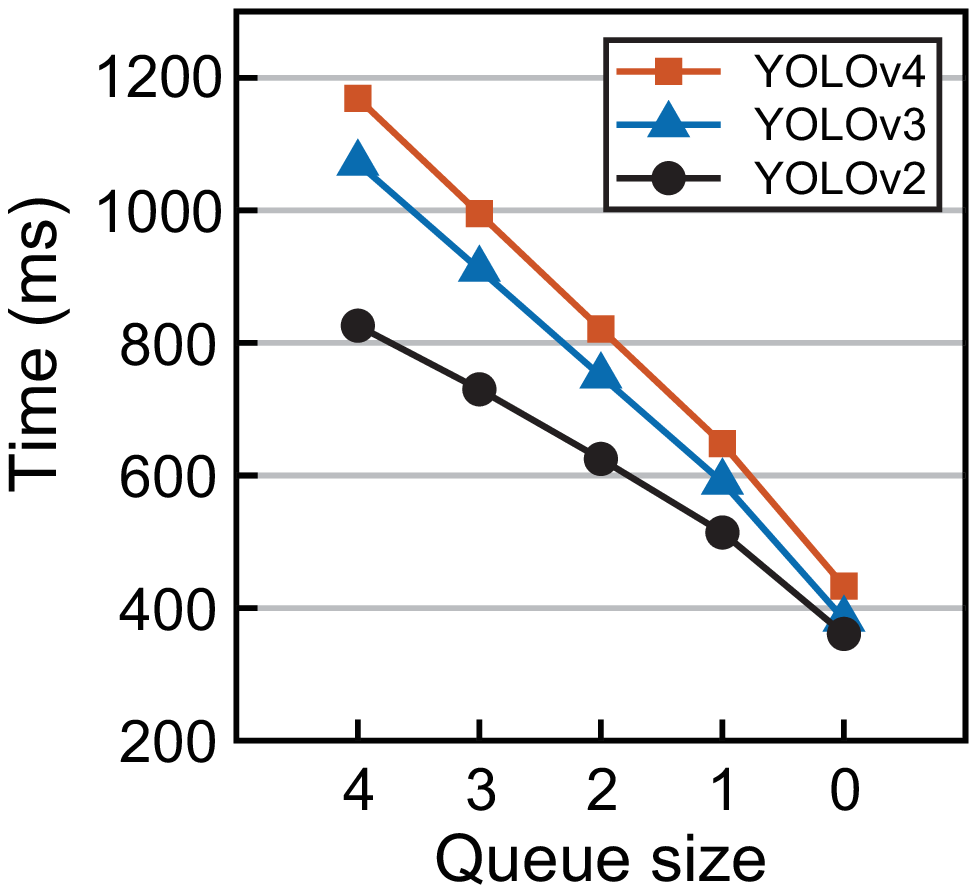}}
\quad
\subfloat[Object detection cycle times]{\label{fig:Qlen_cycle_time}\includegraphics[width=.20\textwidth]{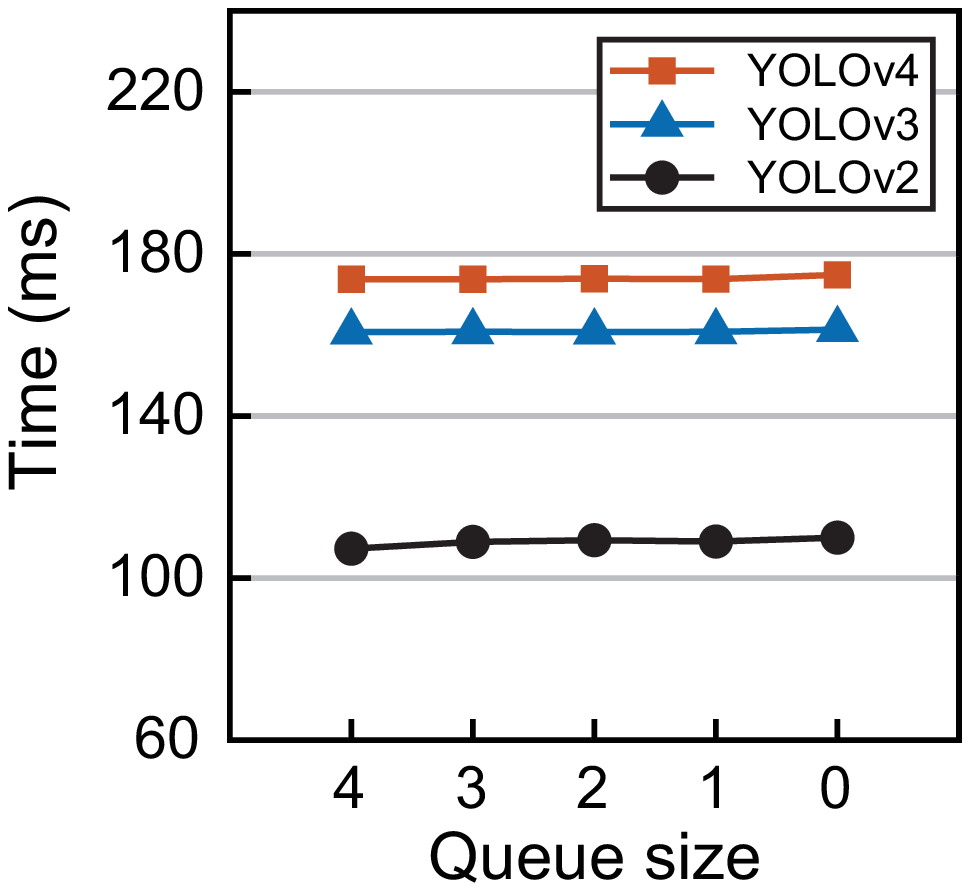}}
\caption{Average end-to-end delays and object detection cycle times with varying queue sizes.}
\label{fig:Qlen}

\end{center}
\vspace{-0.5cm}
\end{figure}

\begin{figure}
\centering{\includegraphics[scale= 0.12]{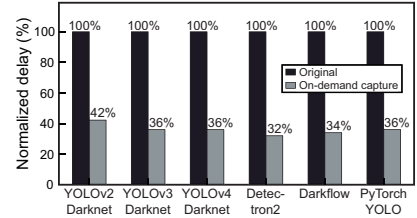}}
\caption{Comparison of normalized average end-to-end delays before and after applying on-demand capture for various object detectors including PyTorch-based YOLO~\cite{pytorch-yolo}.}
\label{fig:sota}
\vspace{-0.5cm}
\end{figure}

This section evaluates our end-to-end delay optimization methods presented in Section~\ref{sec:optimization}. We use the same system configuration as that shown in Table~\ref{tab:configs}, unless otherwise mentioned.

Fig.~\ref{fig:Qlen} shows the end-to-end delays and the object detection cycle times with varying queue sizes for YOLO v2, v3, and v4. In Fig.~\ref{fig:Qlen_delay}, end-to-end delays decrease linearly upon shrinking the queue size, where queue size of zero implies on-demand capture. Further, Fig.~\ref{fig:Qlen_cycle_time} shows that queue sizes do not affect the object detection cycle time. Thus, we can conclude that the queue between the camera subsystem and the detector subsystem is unnecessary, especially when trying to minimize the end-to-end delay.

Since many state-of-the-art object detectors commonly use the OpenCV VideoCapture library with $Q$=4, we tried to replace the original OpenCV library with our modified version anticipating that it could transparently reduce the end-to-end delay of arbitrary object detectors. We did not need to modify the object detectors, because OpenCV is a shared library.  Fig.~\ref{fig:sota} shows how much the end-to-end delays of six different object detectors are reduced by our modified OpenCV library. This figure verifies that on-demand capture is generally applicable to other object detectors.

\begin{figure}
\centering
\subfloat[Average delays]{\label{fig:avg_delay}\includegraphics[width=.24\textwidth]{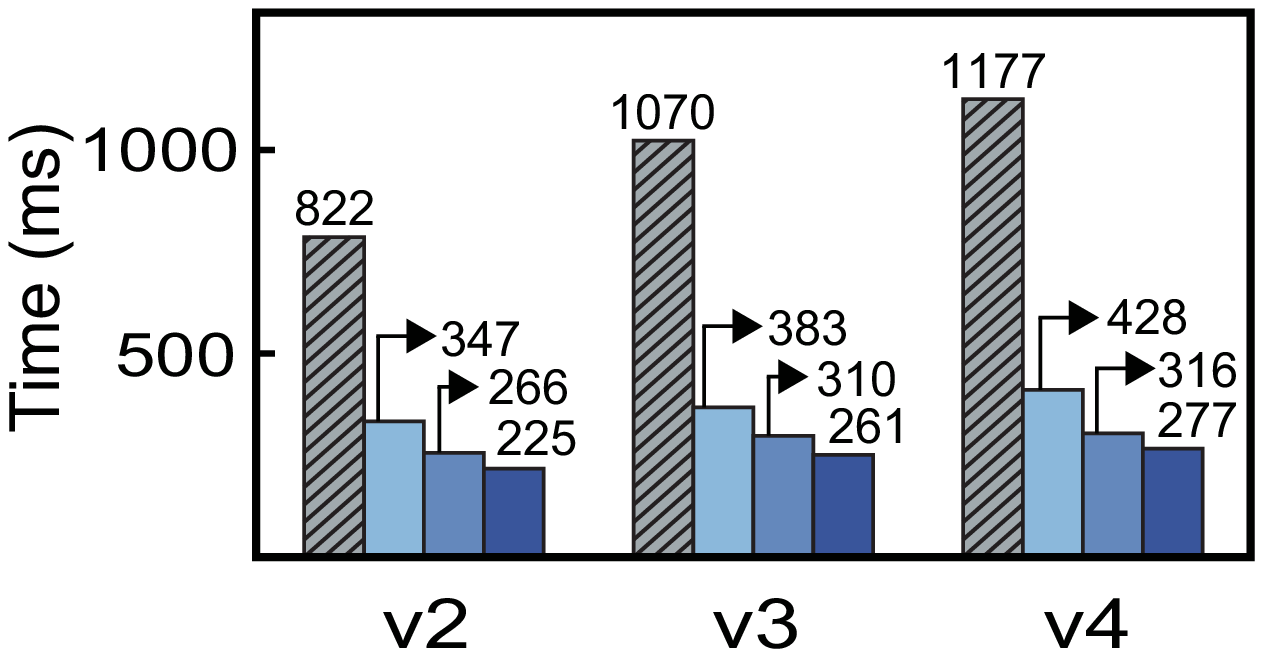}}
\subfloat[Average cycle times]{\label{fig:avg_cycle}\includegraphics[width=.24\textwidth]{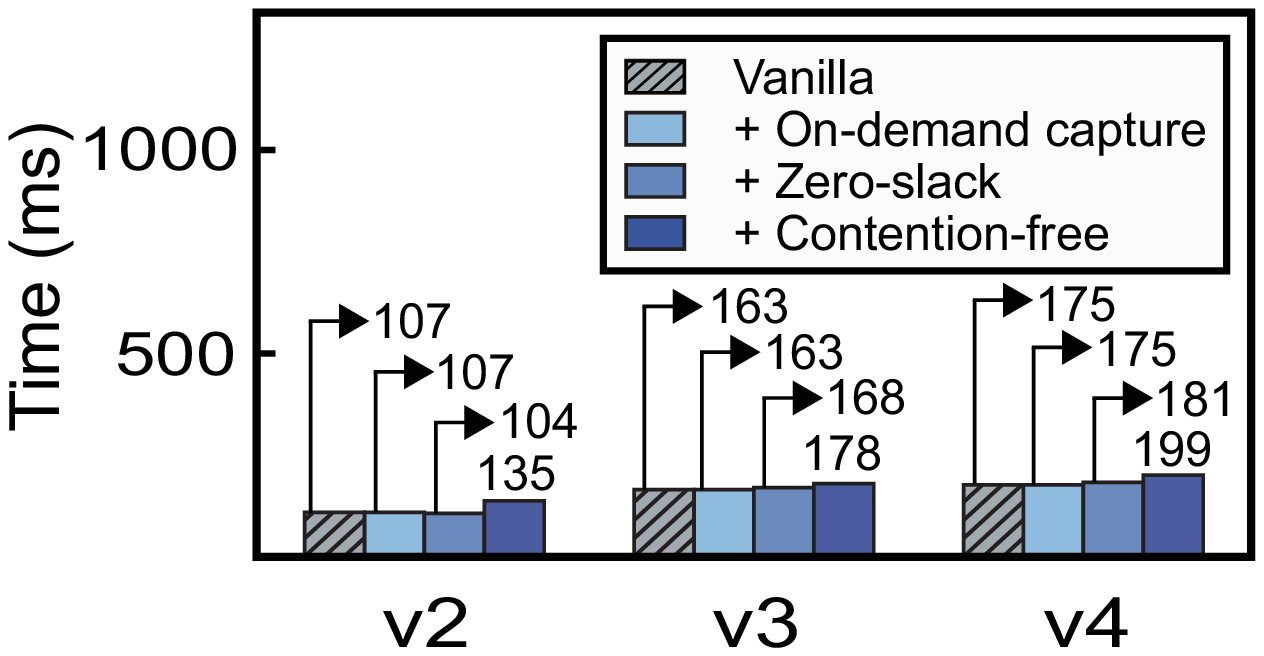}}
\\
\subfloat[99th percentile delays]{\label{fig:99th_delay}\includegraphics[width=.24\textwidth]{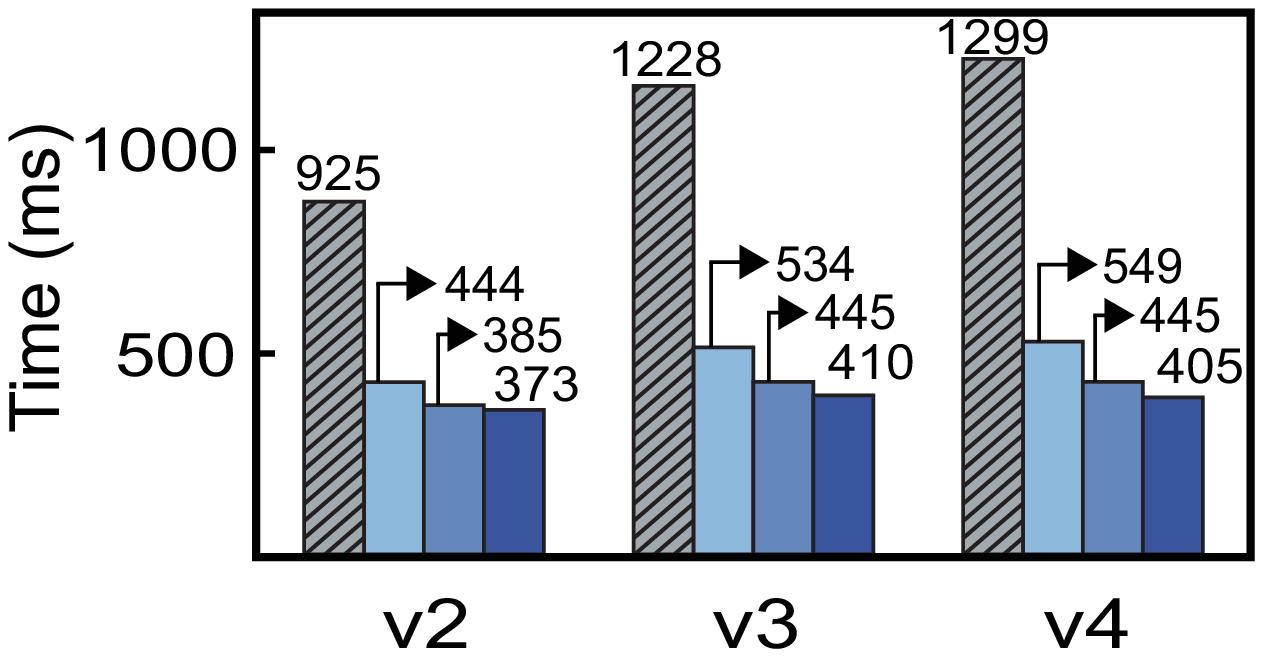}}
\subfloat[99th percentile cycle times]{\label{fig:99th_cycle}\includegraphics[width=.24\textwidth]{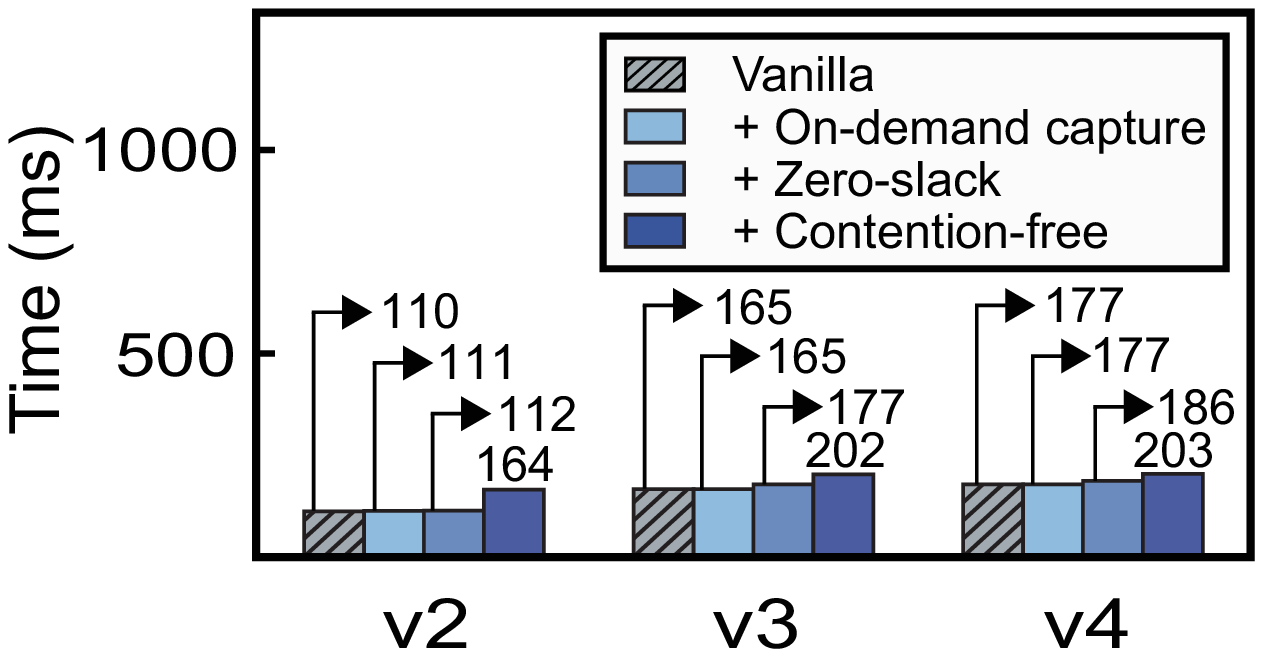}}
\caption{Evaluation results (timing performance).}
\label{fig:eval}
\vspace{-0.5cm}
\end{figure}

Fig.~\ref{fig:eval} shows our evaluation results for the optimization methods. We measured the end-to-end delays and object detection cycle times for YOLO v2, v3, and v4 by incrementally applying the three techniques. For the average system behavior, Figs.~\ref{fig:avg_delay} and \ref{fig:avg_cycle} show the average end-to-end delays and object detection cycle times, respectively. They reveal that the end-to-end delay is continually reduced from the vanilla architecture to the final architecture with the three optimization methods. The average delay reductions are 73\% (v2), 76\% (v3), and 76\% (v4). Regarding the object detection cycle times, on-demand capture and a zero-slack pipeline do not increase the object detection cycle time, whereas a contention-free pipeline slightly increases the object detection cycle time. However, note that the increases are rather marginal. For the tail behavior, Figs.~\ref{fig:99th_delay} and \ref{fig:99th_cycle} show the 99th percentile end-to-end delays and object detection cycle times, respectively: the 99th percentile delays reduced by 60\% (v2), 67\% (v3), and 69\% (v4).

As Darknet has been modified to improve the timing performance, we evaluate its impact on the {\em detection performance} with the KITTI MoSeg dataset~\cite{siam2017modnet,Geiger2013IJRR}. Because we cannot use a physical camera with datasets, we implemented a virtual USB camera by creating a V4L2 loopback device~\cite{v4l2loopback} that reads image files at a constant speed (i.e., 30~fps) and feeds them into the V4L2 queue such that Darknet cannot even notice it is virtual.
\begin{wrapfigure}[9]{r}{.25\textwidth}
  \vspace{-10pt}
    \centering{\includegraphics[width=0.25\textwidth]{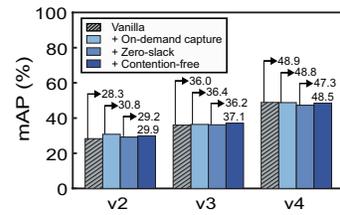}}
  \vspace{-17pt}
  \caption{Detection performance.}
  \label{fig:accuracy}
  \vspace{-0.7cm}
\end{wrapfigure} 
The popular {\em mean average precision (mAP)}~\cite{Everingham15,maplib} is used as the detection accuracy metric. After running 1000 real-world driving images, Fig.~\ref{fig:accuracy} indicates that our approach does not incur any penalty on the detection accuracy. The minute discrepancies are caused by different image capture timings due to the different object detection cycle times. In addition, the trend indicates that more recent YOLO versions show better results.

\section{Related Work}
\label{sec:related}

{\bf Object detection algorithms.}
Recent object detection algorithms are based on DNNs and can be divided into two main categories. One category includes two-stage detectors such as R-CNN~\cite{girshick2014rich}, Fast R-CNN~\cite{girshick2015fast}, Faster R-CNN~\cite{ren2015faster}, and Mask R-CNN~\cite{he2017mask} that use a region proposal network to generate regions of interest in the first stage, where each region of interest is classified in the second stage. The other category includes single-stage detectors such as YOLO, SSD (Single Shot MultiBox Detector)~\cite{liu2016ssd}, and RetinaNet~\cite{lin2017focal} that perform region proposal and classification at the same time. In general, single-stage networks are faster but have slightly lower detection accuracy. Among them, YOLO is widely used in autonomous driving~\cite{lin2018architectural, kato2018autoware, alcon2020timing}. Although this study uses YOLO for the implementation and evaluation, our methods are generally applicable to any of above mentioned object detection algorithms.

{\bf Object detection frameworks.} To train and deploy object detection DNNs, general-purpose DNN frameworks such as TensorFlow~\cite{abadi2016tensorflow}, Caffe~\cite{jia2014caffe}, Caffe2~\cite{caffe2}, and PyTorch~\cite{paszke2019pytorch} can be used. In addition, various specialized object detection frameworks such as Darknet, Darkflow, Detectron~\cite{Detectron2018}, and Detectron2 are available. Among them, we use Darknet as our baseline implementation because it is written purely in C and CUDA, making it portable across different hardware platforms.

{\bf GPU resource management.} To make efficient DNN-based real-time systems, it is important to understand the internal GPU scheduling. However, because this information is not disclosed to the public, many studies~\cite{otterness2016gpu, otterness2017inferring, otterness2017evaluation, amert2017gpu, yang2018avoiding, 8743200} have tried to devise formal GPU scheduling models by empirical methods. To minimize the shared memory bandwidth contention between CPU and GPU, BWLOCK++~\cite{ali2017protecting} protects GPU kernels by performing memory bandwidth throttling, which can be used to augment our method. In \cite{8743176}, a parallel-pipeline execution model for Darknet is presented to improve throughput (i.e., number of cameras) with minor latency increases. On the contrary, our objective is to minimize the end-to-end delay assuming a single camera.


{\bf Real-time DNN inference.} To support DNN inference with real-time requirements, S$^3$DNN~\cite{zhou2018s} provides system-level data fusion and supervised streaming for the efficient utilization of GPUs. A multi-path neural network~\cite{heo2020real} supports dynamic deadlines by managing multiple neural network execution paths with different execution times. In addition, Subflow~\cite{lee2020subflow} supports dynamic deadlines with selective sub-graph executions by taking varying time budgets into account. DART~\cite{9052147} supports the deterministic execution of real-time tasks while providing increased throughput to best-effort tasks. The aforementioned works can be used in conjunction with our method to guarantee explicit real-time requirements in multi-task environments.

\section{Conclusion}
\label{sec:conclusion}

This study was motivated by our observation that many state-of-the-art real-time object detectors show incomprehensibly large end-to-end time lags, despite having decent frame rates. In this light, we provide a complete understanding of the end-to-end delay of an object detection system, in particular, by using Darknet YOLO. 
On the basis of thorough investigation, we developed an end-to-end delay analysis framework that can precisely predict the best- and worst-case end-to-end delays. 
Furthermore, based on the analysis, we presented three optimization methods: (i) on-demand capture, (ii) zero-slack pipeline, and (iii) contention-free pipeline. By applying the above methods, we can reduce the average end-to-end delay of YOLO v3 by 76\% and the 99th percentile delay by 67\% on an Nvidia Jetson AGX Xavier.

In the future, we plan to extend our work with a self-optimizing pipeline framework that automatically reorganizes its pipeline architecture on varying neural networks and hardware platforms, possibly with more complex application scenarios, such as a sensor fusion system with multiple sensors.

\section*{Acknowledgment}

This work was supported partly by the Korea Evaluation Institute Of Industrial Technology (KEIT) grant funded by the Ministry of Trade, Industry and Energy (MOTIE) (20000316, Scene Understanding and Threat Assessment based on Deep Learning for Automatic Emergency Steering), partly by Institute for Information \& communications Technology Promotion (IITP) grant funded by the Korea government (MSIT) (2014-3-00065, Resilient Cyber-Physical Systems Research), partly by the Ministry of Land, Infrastructure, and Transport (MOLIT), Korea, through the Transportation Logistics Development Program (20TLRP-B147674-03, Development of Operation Technology for V2X Truck Platooning), and partially by NSF grant CCF-1704859. J.-C. Kim is the corresponding author of this paper (Co-correspondence: K. Kang).

\bibliographystyle{IEEEtran}
\bibliography{rtod}

\end{document}